\documentclass[nohyperref]{article}

\usepackage{microtype}
\usepackage{graphicx}
\usepackage{subfigure}
\usepackage{booktabs} % for professional tables

\usepackage{hyperref}

\usepackage[accepted]{icml2022}

\usepackage{amsmath}
\usepackage{amssymb}
\usepackage{mathtools}
\usepackage{amsthm}
\usepackage{cancel}
\DeclareMathOperator*{\argmax}{arg\,max}

\usepackage[capitalize,noabbrev]{cleveref}

\theoremstyle{plain}
\newtheorem{theorem}{Theorem}[section]

\newtheorem{lemma}[theorem]{Lemma}
\newtheorem{corollary}[theorem]{Corollary}
\theoremstyle{definition}
\newtheorem{definition}[theorem]{Definition}

\theoremstyle{remark}

\usepackage[textsize=tiny]{todonotes}

\icmltitlerunning{An Analytical Update Rule for General Policy Optimization}

\begin{document}

\twocolumn[
\icmltitle{An Analytical Update Rule for General Policy Optimization}



\icmlsetsymbol{equal}{*}

\begin{icmlauthorlist}
\icmlauthor{Hepeng Li}{yyy}
\icmlauthor{Nicholas Clavette}{yyy}
\icmlauthor{Haibo He}{yyy}
\end{icmlauthorlist}

\icmlaffiliation{yyy}{Department of Electrical, Computer and Biomedical Engineering, University of Rhode Island, South Kingstown, RI, USA}

\icmlcorrespondingauthor{Haibo He}{haibohe@uri.edu}

\icmlkeywords{Machine Learning, ICML}

\vskip 0.3in
]



\printAffiliationsAndNotice{}  % leave blank if no need to mention equal contribution

\begin{abstract}
We present an analytical policy update rule that is independent of parametric function approximators. The policy update rule is suitable for optimizing general stochastic policies and has a monotonic improvement guarantee. It is derived from a closed-form solution to trust-region optimization using calculus of variation, following a new theoretical result that tightens existing bounds for policy improvement using trust-region methods. The update rule builds a connection between policy search methods and value function methods. Moreover, off-policy reinforcement learning algorithms can be derived from the update rule since it does not need to compute integration over on-policy states. In addition, the update rule extends immediately to cooperative multi-agent systems when policy updates are performed by one agent at a time.
\end{abstract}

\section{Introduction}
\label{intro}
Policy search methods have gained great popularity in reinforcement learning (RL) for the last decade. As opposed to value function methods, in which the policy is represented implicitly by a greedy action-selection strategy with respect to an estimated value function, policy search methods search directly in the space of policy representations for a good policy. The advantages of policy search methods include being able to learn stochastic policies \cite{Singh94learningwithout}, better convergence, and effectiveness in high-dimensional or continuous action spaces. Generally, policy search approaches use function approximators, such as neural networks, to construct a parametric policy. The parametric policy is then optimized using policy gradient \cite{10.1007/BF00992696,Sutton1999} or derivative-free algorithms \cite{6796865} by searching in the parameter space.

In this paper, we present an analytical policy update rule that is independent of parametric function approximators. We prove that the update rule has a monotonic improvement guarantee and is suitable for optimizing general stochastic policies with continuous or discrete actions. The update rule provides a new theoretical foundation for policy-based RL, which traditionally restricts the policy search to a family of parametric functions, such as policy gradient \cite{Sutton1999}, deterministic policy gradient \cite{pmlr-v32-silver14, LillicrapHPHETS15}, actor critic \cite{NIPS1999_6449f44a,10.5555/3042573.3042600}, soft actor-critic (SAC) \cite{pmlr-v80-haarnoja18b,sac-arxiv}, and so on.

Our update rule is derived from a closed-form solution to a trust region method using calculus of variation. Trust-region method is one of the most important tools in RL. The basic idea is to search for an improved policy iteratively in a local area around the current policy, in which the objective function is well-approximated by a manageable surrogate model. A representative trust-region method for RL is trust region policy optimization (TRPO) \cite{pmlr-v37-schulman15}. TRPO introduces a simple and functional surrogate model that can be evaluated using the current best policy and provides an upper bound of the approximation error of the surrogate model. This is particularly useful because by subtracting the bound from the surrogate model we obtain the worst-case performance degradation, or a lower bound, of the true objective. It follows that maximizing the lower bound leads to an improved policy with non-decreasing performance \cite{pmlr-v37-schulman15}.

The theory of TRPO is of significance to policy-based RL for it provides an approach that guarantees to improve the policy monotonically. However, the bound derived in TRPO depends on the maximum KL-Divergence of the current policy $\pi$ and a proposed policy $\pi'$ on the entire state space, i.e., $\max_{s}D_{\mathrm{KL}}[\pi'\Vert\pi](s)$, which can be extremely large or infinity even if $\pi'$ and $\pi$ are close at most states. To address this issue, TRPO heuristically imposes a strict constraint to bound the KL-Divergence at every state, but it is intractable to implement this constraint when the state space is large or continuous. To derive a practical algorithm, an empirical approximation using an expected KL-Divergence, e.g. $\mathbb{E}_{s\sim d^{\pi}}[D_{\mathrm{KL}}[\pi'\Vert\pi](s)]$, is usually adopted \cite{pmlr-v37-schulman15,achiam2017constrained}. Nevertheless, the monotonic improvement property is no longer guaranteed.

In this paper, we prove a new theoretical result on the bound of the surrogate approximation error by relating it to the expected KL-Divergence. This result leads to a more practical lower bound of the objective, which improves previous analysis on this topic in terms of KL-Divergence, such as \citet{pmlr-v37-schulman15,achiam2017constrained,10.5555/3291125.3291139}. It also closes the gap between theory and practice in TRPO and the related approaches. Furthermore, this result enables us to derive a closed-form solution for policy optimization. The closed-form solution introduces a very simple policy update rule that guarantees to produce monotonically improving policies.

From an algorithmic viewpoint, the policy update rule enables the development of off-policy algorithms that do not rely on policy gradient \cite{Sutton1999}, which is known to have high variance and low sample efficiency. This is because the policy update rule does not require integrating on-policy distributions over the state space. Thus, we can reuse the past experience obtained from a behavioral policy and circumvent the high variance and sample efficiency issues. In addition, since the policy update rule is analytical, it applies to both parametric and non-parametric policies. However, policy gradient-based approaches are subject to parametric policies.

Furthermore, we prove that the update rule extends immediately to partially observable Markov games with cooperative agents and the monotonic improvement guarantee still holds when updates are performed by one agent at a time.

The contributions of this paper include: (1) a new theoretical result that tightens existing bounds for local policy search using trust-region methods; (2) a closed-form update rule for general stochastic policies with monotonic improvement guarantee; (3) a proof that shows that the policy update rule is extendable to partially observable multi-agent RL problems without compromising the monotonic improvement guarantee.

\section{Related Work}
\label{related_work}
The idea of restricting policy search to a local area of the current policy is common in model-free RL. For instance, instead of imposing a hard boundary on the searching area, \citet{Kakade02approximatelyoptimal} proposed a conservative update scheme mixing the current policy and a greedy update via a weighted sum. A lower bound on the performance improvement as a function of the weighting coefficient was proven. Following this line of work, \citet{pmlr-v28-pirotta13} proposed two more general lower bounds connecting to the difference between two policies. Then, two conservative update algorithms were developed by maximizing the proposed bounds, respectively. \citet{arxiv.2008.10806} proposed a similar bound and a practical algorithm for entropy-regularized RL. While monotonic improvement guarantee is derived in the previous studies, the update scheme cannot apply to non-mixture policies. \citet{pmlr-v37-schulman15} extended this line of work to general stochastic policies and proposed a new bound that connected it to the maximum KL-Divergence between two successive policies on the state space. However, this bound is intractable when the state space is large. Although a tighter bound relating it to an average total variation distance is proposed in \citet{achiam2017constrained}, deriving a closed-form policy update rule from the lower bound is still challenging.

In practice, many approaches use a hard constraint to bound the searching area but they generally lose the monotonic improvement guarantee. \citet{10.5555/2898607.2898863} proposed relative entropy policy search (REPS) to restrict the relative entropy between observed data distribution of the state-action pairs and the distribution generated by the new policy. A closed-form update rule in a softmax form was derived using the method of Lagrange multipliers. However, this approach is not straightforwardly extendable to general non-linear policies. To apply nonlinear policies, TRPO \cite{pmlr-v37-schulman15} and constrained policy optimization \cite{achiam2017constrained} approximately constrained the on-policy expected KL-Divergence by using second-order Taylor expansion, which was closely related to natural policy gradient \cite{10.5555/2980539.2980738}. Extending the work in TRPO, \citet{10.5555/3291125.3291139} provided a monotonic improvement guarantee for bounding the expected KL-divergence, but the result only held for linear-Gaussian policies. \citet{10.5555/3294996.3295037,nachum2018trustpcl} presented multi-step softmax consistencies under entropy regularization and adopted a discounted relative entropy trust-region constraint to improve exploration and stability. By relating policy search to probabilistic inference \cite{levine2018reinforcement}, \citet{abdolmaleki2018maximum} proposed the maximum a posterior policy optimization (MPO) algorithm based on Expectation-Maximization, where the policy update was decomposed into E-step and M-step. A closed-form E-step combined with a maximum-a-posteriori-estimation M-step for Gaussian policies was provided. Although a monotonic improvement guarantee is claimed, the guarantee is for the KL-Divergence regularized objective rather than the true expected return. Besides, suffering from the same issue as in \cite{10.5555/2898607.2898863}, the policy update rule needs to determine the optimal Lagrangian multipliers of the dual problem, which requires a costly nonlinear optimization in the inner loop. Different from previous works, \citet{otto2021differentiable} proposed projection-based solutions to impose trust-region constraints on the individual state, which enabled exact guarantees of monotonic improvement. Three closed-form projection layers based on Wasserstein L2 distance, Frobenius norm, and KL-Divergence were proposed to project the updated policy onto trust regions. However, the proposed approach only applies to Gaussian policies.
\section{Preliminaries}
\label{preliminaries}
\subsection{Markov Decision Process}
A Markov decision process (MDP) is defined by a tuple $(\mathcal{S},\mathcal{A},p,r,\rho_{0},\gamma)$, where $\mathcal{S}$ is the state space, $\mathcal{A}$ is the action space, $p:\mathcal{S}\times\mathcal{A}\times\mathcal{S}\rightarrow\mathbb{R}_{\geq 0}$ is the transition probability density, $r:\mathcal{S}\times\mathcal{A}\rightarrow [r_{\mathrm{min}},r_{\mathrm{max}}]$ is the reward function, $\rho_{0}:\mathcal{S}\rightarrow\mathbb{R}_{\geq 0}$ is the probability density of the initial state $s_{0}$, $\gamma\in[0,1)$ is the discount factor.

Denote a stochastic policy $\pi:\mathcal{S}\times\mathcal{A}\rightarrow\mathbb{R}_{\geq 0}$ ($\rightarrow [0,1]$ for discrete actions) by $\pi(a|s)$, which represents the probability density (or probability mass function) of the action $a$ given the state $s$.
The goal is to find an optimal policy that maximizes the expected discounted return
\begin{equation}
\begin{split}
	&J(\pi)=\mathbb{E}_{\tau\sim\pi}\left[\sum_{t=0}^{\infty}\gamma^{t}r_{t}\right]
\end{split}
\end{equation}
where $\tau$ denotes the trajectory $\tau:=(s_{0},a_{0},s_{1},\dots)$, and $\tau\sim\pi$ indicates that the distribution over the trajectory depends on $\pi:s_{0}\sim\rho_{0},a_{t}\sim\pi(\cdot|s_{t}),s_{t+1}\sim p(\cdot|s_{t}, a_{t})$. Letting $R(\tau)=\sum_{t=0}^{\infty}\gamma^{t}r_{t}$ denote the discounted return of the trajectory $\tau$, we can compactly express the value function as $V_{\pi}(s)=\mathbb{E}_{\tau\sim\pi}\left[R(\tau)|s_{0}=s\right]$, the state-action value function as $Q_{\pi}(s,a)=\mathbb{E}_{\tau\sim\pi}\left[R(\tau)|s_{0}=s,a_{0}=a\right]$, and the advantage function as $A_{\pi}(s, a)=Q_{\pi}(s,a)-V_{\pi}(s)$. We define the discounted state visitation distribution as
\begin{equation}
	d^{\pi}(s)=(1-\gamma)\sum_{t=0}^{\infty}\gamma^{t}\rho^{\pi}_{t}(s),
\end{equation}
where $\rho^{\pi}_{t}:\mathcal{S}\rightarrow\mathbb{R}_{\geq 0}$ is probability density function (PDF) of the state at timestep $t$ given the policy $\pi$.

\subsection{Partially Observable Markov Game}
A Markov game \cite{Michael1994} is a game defined on a state space, $\mathcal{S}$, and a collection of action spaces, $\mathcal{A}^{1},\dots,\mathcal{A}^{N}$, one for each agent in the environment. The state transition $s\mapsto s^{'} (s,s^{'}\in\mathcal{S})$ happens following the probability density $P:\mathcal{S}\times\mathcal{A}^{1}\times\cdots\times\mathcal{A}^{N}\times\mathcal{S}\mapsto\mathbb{R}_{\geq 0}$ when the actions $a=[a^{1},\dots,a^{N}]$, $a^{i}\in\mathcal{A}^{i}$, $i\in\{1,\dots,N\}$, are exerted on the environment at state $s$. Each agent is rewarded based on a local reward function $r^{i}:\mathcal{S}\times\mathcal{A}^{1}\times\cdots\times\mathcal{A}^{N}\mapsto[r^{i}_{\mathrm{min}},r^{i}_{\mathrm{max}}]$, which depends on the current state $s$ and the joint action $a$.

In a partially observable Markov game (POMG), each agent has a local observation of the environment, $o^{i}$, which contains incomplete information of the state $s$. At state $s$, $o^{i}$ is observed with a likelihood, $P_{o}^{i}:\mathcal{S}\times\mathcal{O}^{i}\mapsto\mathbb{R}_{\geq 0}$, where $\mathcal{O}^{i}$ is the observation space of the agent. Each agent acts according to a policy $\pi^{i}:\mathcal{O}^{i}\times\mathcal{A}^{i}\mapsto\mathbb{R}_{\geq 0}\ (\text{or} \mapsto[0,1])$, which is a probability distribution (or a probability mass function) over the action space $\mathcal{A}^{i}$ given the observation $o^{i}$. We will use the following definitions of the joint policy $\pi(a|s)$ and the joint policy $\pi^{-i}(a^{-i}|s)$ except $i$:
\begin{equation}
\label{joint_policy}
	\pi(a|s)=\prod_{i\in\mathcal{N}}\int_{o^{i}}\pi^{i}(a^{i}|o^{i})P_{o}^{i}(o^{i}|s)do^{i}
\end{equation}
\begin{equation}
\label{others_policy}
	\pi^{-i}(a^{-i}|s)=\prod_{j\in\mathcal{N}\setminus\{i\}}\int_{o^{j}}\pi^{j}(a^{j}|o^{j})P_{o}^{j}(o^{j}|s)do^{j}
\end{equation}
where $\mathcal{N}=\{1,2,\dots,N\}$ is a set of agent's IDs. The goal of the agents is to learn a set of distributed policies $\{\pi^{i}(a^{i}|o^{i})|i\in\mathcal{N}\}$ to maximize the expected return
\begin{equation}
\begin{split}
	&J(\pi)=\mathbb{E}_{\tau\sim\pi}\left[\sum_{t=0}^{\infty}\gamma^{t}(r_{t}^{1}+\cdots+r_{t}^{N})\right]
\end{split}
\end{equation}
where $\tau\sim\pi$ indicates that $s_{0}\sim\rho_{0},o_{t}^{i}\sim P_{o}^{i}(\cdot|s_{t}),a_{t}^{i}\sim\pi^{i}(\cdot|o_{t}^{i}),s_{t+1}\sim P(\cdot|s_{t},a_{t}^{1},...,a_{t}^{N})$.
\subsection{Trust Region Method}
Trust region method is one of the most important techniques for solving policy optimization in a Markov decision process. It works by restricting policy search to a local region around the current best solution, where the objective function is well-approximated by a surrogate model. Specifically, it solves the following optimization:
\begin{equation}
	\pi_{k+1}=\argmax_{\pi'\in\Pi} \tilde{J}(\pi'),\ s.t.\ D(\pi',\pi_{k})\leq\delta
\end{equation}
where $\tilde{J}(\pi')$ is some surrogate model, $D$ is a distance measure, and $\delta>0$ is the radius of a spherical region, in which we search for an improved policy. A simple and effective choice for the surrogate model is 
\begin{equation}
\label{surrogate}
	L_{\pi_{k}}(\pi')=J(\pi_{k})+\frac{1}{1-\gamma}\mathbb{E}_{s\sim d^{\pi_{k}}, a\sim\pi'}[A_{\pi_{k}}(s,a)].
\end{equation}
\citet{pmlr-v37-schulman15} prove that the difference between the surrogate model and the true objective is bounded by:
\begin{equation}
\label{bound}
\begin{aligned}
	&\big\vert J(\pi') - L_{\pi_{k}}(\pi')\big\vert \leq C\max_{s}D_{\mathrm{KL}}[\pi'\Vert\pi_{k}](s), \\
	&\text{\emph{where}}\ C=\frac{4\gamma\epsilon}{(1-\gamma)^2},\ \epsilon=\max_{s,a}|A_{\pi_{k}}(s,a)|,
\end{aligned}
\end{equation}
which connects it to the maximum KL-Divergence over the state space, $\max_{s}D_{\mathrm{KL}}[\pi'\Vert\pi_{k}](s)$. By using this bound, we can get the worst-case performance degradation of the true objective:
\begin{equation}
\label{lower_bound}
	J(\pi')\geq L_{\pi_{k}}(\pi')-C\max_{s}D_{\mathrm{KL}}[\pi'\Vert\pi_{k}](s), 
\end{equation}
It follows that maximizing the right-hand side of the inequality, which is a lower bound of the true objective function, can lead to guaranteed improvement in the performance. This result has fostered a branch of practical trust-region algorithms (i.e. \citet{pmlr-v37-schulman15,schulman2017proximal,achiam2017constrained,nachum2018trustpcl,10.5555/3295222.3295280}) that approximately optimize the lower bound to improve policies. 
\section{Analytical Policy Update Rule with Monotonic Improvement Guarantee}
Our principle result is an analytical solution for policy optimization based on trust-region  methods, following a new bound on the difference between the surrogate model and the true objective. The analytical solution introduces a policy update rule that guarantees monotonic policy improvement and is suitable for general stochastic policies with discrete or continuous actions. Moreover, the update rule extends immediately to cooperative multi-agent systems when updates are performed by one agent at a time.

We first present the new bound on the difference between the surrogate model and the objective in the following theorem.
\begin{theorem}
\label{new_bound}
For any stochastic policies $\pi',\pi$ and discount factor $\gamma\in[0.5,1)$, the following bound holds:
\begin{equation}
\begin{aligned}
	&\big\vert J(\pi') - L_{\pi}(\pi')\big\vert \leq \frac{1}{1-\gamma}C_{\pi}\mathbb{E}_{s\sim d^{\pi}}\left[D_{\mathrm{KL}}[\pi'\Vert\pi](s)\right], \\
	&\text{where}\ C_{\pi}=\frac{\gamma^{2}\epsilon}{(1-\gamma)^{3}},\ \epsilon=\max_{s,a}|A_{\pi}(s,a)|.
\end{aligned}
\end{equation}
\end{theorem}
\begin{proof}[Proof]
We provide the proof in Appendix \ref{appendix-A}. The proof extends \citet{pmlr-v37-schulman15}'s result using the concept of $\alpha$-coupling \cite{LevinPeresWilmer2006} and its relationship with total variation distance. However, different from the proof in \cite{pmlr-v37-schulman15} that uses the maximum $\alpha$ over the state space, we instead use a state-dependent $\alpha(s)$ to represent the coupling between two arbitrary policies given $s$, which enables us to connect the bound to the expected KL-Divergence.
\end{proof}
The new bound is tighter in terms of KL-Divergence compared with \eqref{bound} derived from \cite{pmlr-v37-schulman15}. While the improvement in tightness is at a cost of $\gamma/(4(1-\gamma)^2)$, this result directly relates the bound to the expected KL-Divergence $\mathbb{E}_{s\sim d^{\pi}}\left[D_{\mathrm{KL}}[\pi'\Vert\pi](s)\right]$, which closes the gap between theory and practice in TRPO and related algorithms. In addition, the new bound improves prior analysis in the literature, such as \cite{10.5555/3291125.3291139,achiam2017constrained}, in terms of either KL-Divergence or total variation distance (from $D_{\mathrm{TV}}[\pi'\Vert\pi]$ to $D_{\mathrm{TV}}^{2}[\pi'\Vert\pi]$, see Appendix \ref{appendix-A}). Furthermore, using this result, we can derive a new lower bound of the true objective:
\begin{equation}
\label{new_lower_bound}
	J(\pi')\geq L_{\pi_{k}}(\pi')-\frac{1}{1-\gamma}C_{\pi}\mathbb{E}_{s\sim d^{\pi}}\left[D_{\mathrm{KL}}[\pi'\Vert\pi](s)\right].
\end{equation}
Then, we can improve the policy by maximizing the lower bound. Next, we present a closed-form solution to the maximization of the lower bound, which introduces a simple policy update rule with monotonic improvement guarantee.
\begin{theorem}
\label{formula}
For any stochastic policies $\pi_{\mathrm{new}},\pi_{\mathrm{old}}$ that are continuously differentiable on the state space $\mathcal{S}$, the inequality, $J(\pi_{\mathrm{new}})\geq J(\pi_{\mathrm{old}})$, holds when
\begin{equation}
\label{update_formula}
\begin{aligned}
\pi_{\mathrm{new}}=\pi_{\mathrm{old}}\cdot\frac{e^{\alpha_{\pi_{\mathrm{old}}}}}{\mathbb{E}_{a\sim\pi_{\mathrm{old}}}\left[e^{\alpha_{\pi_{\mathrm{old}}}}\right]},
\end{aligned}
\end{equation}
where $\alpha_{\pi_{\mathrm{old}}}=A_{\pi_{\mathrm{old}}}/C_{\pi_{\mathrm{old}}}$.
\end{theorem}
\begin{proof}[Proof]
We provide the proof in Appendix \ref{appendix-B}. The proof introduces calculus of variation \cite{noauthororeditor,MarkKot2014} to the policy optimization problem. Based on the assumption of continuously differentiable policies on the state space $\mathcal{S}$, we derive a closed-form solution for general stochastic policies with continuous or discrete actions. In the proof, we show that the closed-form solution is a necessary and sufficient condition for the policy optimization.
\end{proof}
Another interesting result of Theorem \ref{formula} is that the update rule immediately extends to cooperative multi-agent RL problems while the monotonic improvement guarantee still holds if the agents perform local policy updates in turn. We present this result in the following corollary.
\begin{corollary}
For any stochastic policies $\pi_{\mathrm{new}}^{i},\pi_{\mathrm{old}}^{i}$ of agent $i$ that are continuously differentiable on the local observation space $\mathcal{O}^{i}$, and the corresponding joint policies $\pi_{\mathrm{new}},\pi_{\mathrm{old}}$, the inequality, $J(\pi_{\mathrm{new}})\geq J(\pi_{\mathrm{old}})$, holds when
\begin{equation}
\begin{split}
	&\pi_{\mathrm{new}}^{i}=\pi_{\mathrm{old}}^{i}\cdot\frac{e^{\alpha_{\pi_{\mathrm{old}}}}}{\mathbb{E}_{a\sim\pi_{\mathrm{old}}}\left[e^{\alpha_{\pi_{\mathrm{old}}}}\right]}, \\
	&\pi_{\mathrm{new}}^{-i}=\pi_{\mathrm{old}}^{-i}.
\end{split}
\end{equation}
where $\pi_{\mathrm{new}}^{-i},\pi_{\mathrm{old}}^{-i}$ are joint policies of all agents except $i$.
\end{corollary}
\begin{proof}[Proof]
Based on Theorem \ref{formula}, we have
\begin{equation}
\label{proof_update_formula_multiagent}
\pi_{\mathrm{new}}(a|s)=\pi_{\mathrm{old}}(a|s)\frac{e^{\alpha_{\pi_{\mathrm{old}}}}}{\mathbb{E}_{a\sim\pi_{\mathrm{old}}}\left[e^{\alpha_{\pi_{\mathrm{old}}}}\right]}\end{equation}

Note that the joint policy can be decomposed as follows:
\begin{equation}
	\pi(a|s)=\pi^{i}(a^{i}|s)\pi^{-i}(a^{-i}|s),
\end{equation}
where $\pi^{i}(a^{i}|s)=\int_{o^{i}}\pi^{i}(a^{i}|o^{i})P_{o}^{i}(o^{i}|s)do^{i}$. Thus, we can rewrite Eq. (\ref{proof_update_formula_multiagent}) as follows:
\begin{equation}
\label{ma_update_rule}
\begin{split}
	&\ \pi_{\mathrm{new}}^{-i}\int_{\mathcal{O}^{i}}\pi_{\mathrm{new}}^{i}(a^{i}|o^{i})P_{o}^{i}(o^{i}|s)do^{i} \\
	=&\ \pi_{\mathrm{old}}^{-i}\int_{\mathcal{O}^{i}}\pi_{\mathrm{old}}^{i}(a^{i}|o^{i})P_{o}^{i}(o^{i}|s)do^{i}\cdot\frac{e^{\alpha_{\pi_{\mathrm{old}}}}}{\mathbb{E}_{a\sim\pi_{\mathrm{old}}}\left[e^{\alpha_{\pi_{\mathrm{old}}}}\right]},
\end{split}
\end{equation}
when $\pi_{\mathrm{new}}^{-i}=\pi_{\mathrm{old}}^{-i}$, they cancel each other on both sides. Then, simplifying the above equation, the result follows.
\end{proof}
\section{Connections with Prior Work}
In this section, we connect the proposed policy update rule with some state-of-the-art algorithms and discuss how the update rule can help explain these algorithms from a different perspective.
\subsection{TRPO and Proximal Policy Optimization}
\begin{figure}[t!]
\centering     %%% not \center
\subfigure{\includegraphics[width=82mm]{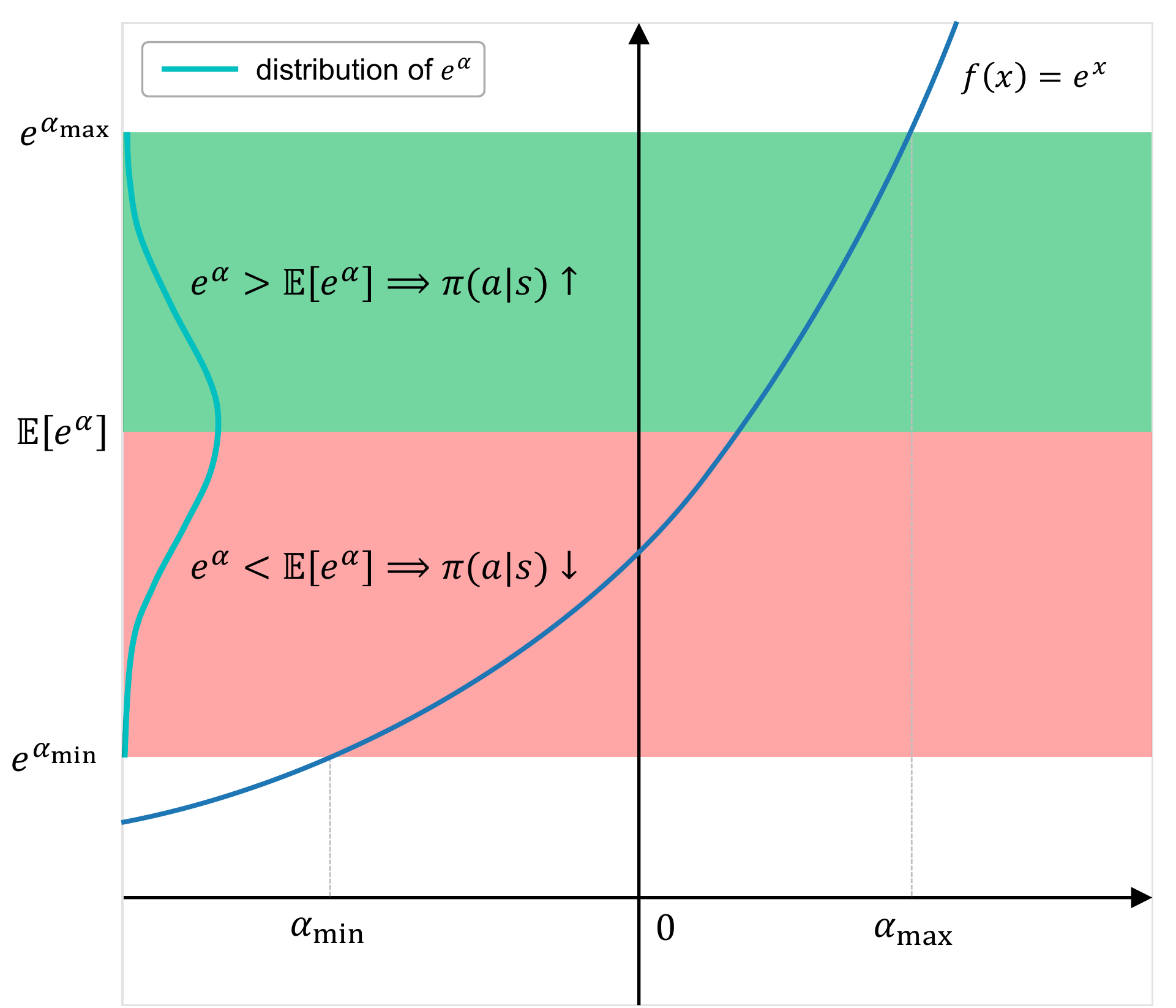}}
\caption[Caption for LOF]{From the policy update rule, we can derive $\pi_{\mathrm{new}}/\pi_{\mathrm{old}}\in\left[e^{\alpha_{\mathrm{min}}}/Z,e^{\alpha_{\mathrm{max}}}/Z\right]$, where $Z=\mathbb{E}_{a\sim\pi_{\mathrm{old}}}\left[e^{\alpha_{\pi_{\mathrm{old}}}}\right]$ and $\alpha_{\mathrm{min}}$ and $\alpha_{\mathrm{max}}$ are the minimum and maximum values of $\alpha_{\pi_{\mathrm{old}}}$, respectively. Since $e^{\alpha_{\mathrm{min}}}\leq \mathbb{E}_{a\sim\pi_{\mathrm{old}}}\left[e^{\alpha_{\pi_{\mathrm{old}}}}\right]\leq e^{\alpha_{\mathrm{max}}}$, the upper and lower bounds of the policy ratio $\pi_{\mathrm{new}}/\pi_{\mathrm{old}}$ can be expressed as $[1-\epsilon_{1},1+\epsilon_{2}]$, where $\epsilon_{1},\epsilon_{2}\geq 0$ and $\epsilon_{1}<1$. This indicates that we can bound the policy ratio to restrict the search area, which has been adopted in PPO and proven effective in practice. In addition, if $e^{\alpha_{\pi_{\mathrm{old}}}}>\mathbb{E}_{a\sim\pi_{\mathrm{old}}}\left[e^{\alpha_{\pi_{\mathrm{old}}}}\right]$, the policy ratio $\pi_{\mathrm{new}}/\pi_{\mathrm{old}}$ will be greater than 1, and vice versa. Note that $e^{\alpha_{\pi_{\mathrm{old}}}}$ is monotonically increasing with respect to $A_{\pi_{\mathrm{old}}}(s,a)$, and so is the ratio $\pi_{\mathrm{new}}/\pi_{\mathrm{old}}$. Less rigorously, consider the term $\mathbb{E}_{a\sim\pi_{\mathrm{old}}}\left[e^{\alpha_{\pi_{\mathrm{old}}}}\right]$ as the ``average'' advantage of the policy $\pi_{\mathrm{old}}$. Then, selecting the action $a$ at state $s$ is encouraged, i.e. $\pi_{\mathrm{new}}(a|s)>\pi_{\mathrm{old}}(a|s)$, if it leads to an ``above average'' advantage. On the contrary, selecting the action $a$ at state $s$ is discouraged, i.e. $\pi_{\mathrm{new}}(a|s)<\pi_{\mathrm{old}}(a|s)$, if it leads to a ``below average'' advantage. To what extent the action $a$ is encouraged or discouraged is determined by the value of $A_{\pi_{\mathrm{old}}}(s,a)$.}%\footnotemark.}
\label{explanation}
\end{figure}

Note that the exponential factor in (\ref{update_formula}) can be written as
\begin{equation}
\label{AoverC}
	\alpha_{\pi_{\mathrm{old}}}=\frac{A_{\pi_{\mathrm{old}}}(s,a)}{\max_{s,a}|A_{\pi_{\mathrm{old}}}(s,a)|}\cdot\frac{(1-\gamma)^{3}}{\gamma^{2}},
\end{equation}
where $\gamma\in[0.5,1)$. The first term on the right-hand side is a normalized advantage and the second term is a positive constant smaller than 1. Letting $[\alpha_{\mathrm{min}}, \alpha_{\mathrm{max}}]$ denote the range of $\alpha_{\pi_{\mathrm{old}}}$, then we have $\alpha_{\mathrm{min}}\leq 0 \leq \alpha_{\mathrm{max}}$, as shown in Figure \ref{explanation}. In addition, since $\alpha_{\pi_{\mathrm{old}}}$ is a random variable given $s$, we have $e^{\alpha_{\mathrm{min}}}\leq \mathbb{E}_{a\sim\pi_{\mathrm{old}}}\left[e^{\alpha_{\pi_{\mathrm{old}}}}\right] \leq e^{\alpha_{\mathrm{max}}}$. Then, based on the update rule \eqref{update_formula}, the ratio of the new policy to the old policy is bounded by
\begin{equation}
\label{ratio}
	\frac{\pi_{\mathrm{new}}}{\pi_{\mathrm{old}}}\in\left[\frac{e^{\alpha_{\mathrm{min}}}}{Z},\frac{e^{\alpha_{\mathrm{max}}}}{Z}\right]=[1-\epsilon_{1}, 1+\epsilon_{2}],
\end{equation}
where $Z=\mathbb{E}_{a\sim\pi_{\mathrm{old}}}\left[e^{\alpha_{\pi_{\mathrm{old}}}}\right]$ and $\epsilon_{1},\epsilon_{2}$ are positive numbers ($\epsilon_{1}<1$). Equation (\ref{ratio}) indicates that bounding the policy ratio is an effective way to confine the searching area. This help explain the success of the proximal policy optimization (PPO) algorithm \cite{schulman2017proximal}, which clips the policy ratio by $[1-\epsilon, 1+\epsilon], 0<\epsilon<1$.

It is also noted that the policy ratio $\pi_{\mathrm{new}}/\pi_{\mathrm{old}}$ will be greater than 1 if $e^{\alpha_{\pi_{\mathrm{old}}}}>\mathbb{E}_{a\sim\pi_{\mathrm{old}}}\left[e^{\alpha_{\pi_{\mathrm{old}}}}\right]$, and vice versa (shown in Figure \ref{explanation}). Note that the exponential term $e^{\alpha_{\pi_{\mathrm{old}}}}$ is monotonically increasing with respect to $A_{\pi_{\mathrm{old}}}(s,a)$, and so is the policy ratio $\pi_{\mathrm{new}}/\pi_{\mathrm{old}}$. Less rigorously, consider the term $\mathbb{E}_{a\sim\pi_{\mathrm{old}}}\left[e^{\alpha_{\pi_{\mathrm{old}}}}\right]$ as an ``average'' advantage of the policy $\pi_{\mathrm{old}}$. Then, selecting the action $a$ at state $s$ is encouraged, i.e. $\pi_{\mathrm{new}}(a|s)>\pi_{\mathrm{old}}(a|s)$, if it leads to an advantage that is above average. On the contrary, selecting the action $a$ at state $s$ is discouraged, i.e. $\pi_{\mathrm{new}}(a|s)<\pi_{\mathrm{old}}(a|s)$, if it leads to an advantage that is below average. To what extent the action $a$ is encouraged or discouraged is determined by the value of $A_{\pi_{\mathrm{old}}}(s,a)$. This result matches the TRPO algorithm \cite{pmlr-v37-schulman15}, which maximizes
\begin{equation}
\begin{split}
	\max_{\pi}\mathbb{E}_{\substack{s\sim d^{\pi_{\mathrm{old}}},a\sim\pi_{\mathrm{old}}}}\left[\frac{\pi(a|s)}{\pi_{\mathrm{old}}(a|s)}A_{\pi_{old}}(s,a)\right],
\end{split}
\end{equation}
where $\pi(a|s)$ is increased to gain weights for large advantages and decreased to lose weights for small advantages. Although our update rule suggests that the policy ratio is proportional to an \emph{exponential} advantage, rather than a \emph{linear} advantage as suggested in TRPO and PPO, it is easy to verify that $e^{A_{\mathrm{old}}/C_{\mathrm{old}}}\approx A_{\mathrm{old}}/C_{\mathrm{old}}+1$ when the policy ratio is bounded around 1.
\subsection{Value-Based Methods and Dynamic Programming}
In this section, we provide a different explanation of the policy update rule by considering discrete actions and then connect it to value function methods. By multiplying the numerator and denominator both by $e^{V_{\pi_{\mathrm{old}}}(s)/C_{\pi_{\mathrm{old}}}}$, we can rewrite the update rule as
\begin{equation}
\label{weighted_explanation}
\begin{aligned}
	&\pi_{\mathrm{new}}(a^{i}|s)=\frac{\pi_{\mathrm{old}}(a^{i}|s)\omega^{i}_{\mathrm{old}}}{\sum_{j}\pi_{\mathrm{old}}(a^{j}|s)\omega^{j}_{\mathrm{old}}},\ \ \ \\
	&\omega^{i}_{\mathrm{old}}=\exp\{Q_{\pi_{\mathrm{old}}}(s,a^{i})/C_{\pi_{\mathrm{old}}}\}\ \ \ 
\end{aligned}
\end{equation}
As shown in (\ref{weighted_explanation}), the new policy is a weighted probability mass function of the old policy in a softmax form. The weights are the exponential terms, $\exp\{Q_{\pi_{\mathrm{old}}}(s,a^{i})/C_{\pi_{\mathrm{old}}}\}$. That indicates actions with larger Q values will get better chance to be selected in the future. In fact, the policy update rule can be deemed as a stochastic analogy of the $\epsilon$-greedy policy used in value function methods, such as SARSA \cite{sutton2018reinforcement}.

In addition, we can verify the monotonic improvement guarantee of the policy update rule via dynamic programming. To see this, we will show $V_{\pi_{\mathrm{old}}}(s)\leq V_{\pi_{\mathrm{new}}}(s)$ for all $s\in\mathcal{S}$. Note that
\begin{equation}
\begin{aligned}
	V_{\pi_{\mathrm{old}}}(s) =&\ \sum_{i}\pi_{\mathrm{old}}(a^{i}|s)Q_{\pi_{\mathrm{old}}}(s,a^{i}) \\
	\leq&\ \sum_{i}\frac{\pi_{\mathrm{old}}(a^{i}|s)\omega^{i}_{\mathrm{old}}}{\sum_{j}\pi_{\mathrm{old}}(a^{j}|s)\omega^{j}_{\mathrm{old}}}Q_{\pi_{\mathrm{old}}}(s,a^{i}) \\
	=&\ \sum_{i}\pi_{\mathrm{new}}(a^{i}|s)Q_{\pi_{\mathrm{old}}}(s,a^{i}).
\end{aligned}
\end{equation}
For brevity, we will use $P_{sa}^{s'}:=P(s'|s,a)$. Then, we have
\begin{equation}
\begin{aligned}
	V_{\pi_{\mathrm{old}}}(s) \leq&\ \mathbb{E}_{\pi_{\mathrm{new}}}\left[Q_{\pi_{\mathrm{old}}}(s,a)\right] \\
	=&\ \mathbb{E}_{\pi_{\mathrm{new}}}\left[r(s,a)+\gamma\mathbb{E}_{P_{sa}^{s'}}[V_{\pi_{\mathrm{old}}}(s')]\right] \\
	\leq&\ \mathbb{E}_{\pi_{\mathrm{new}}}\left[r(s,a)+\gamma\mathbb{E}_{P_{sa}^{s'}}\big[\mathbb{E}_{\pi_{\mathrm{new}}}[Q_{\pi_{\mathrm{old}}}(s',a')]\big]\right] \\
	&\ \vdots \\
	\leq&\ \mathbb{E}_{\pi_{\mathrm{new}}}\left[r(s,a)+\gamma\mathbb{E}_{P_{sa}^{s'}}[r(s',a')]+\cdots\right] \\
	=&\ V_{\pi_{\mathrm{new}}}(s).
\end{aligned}
\end{equation}
Therefore, by applying the update rule (\ref{weighted_explanation}), we can obtain a sequence of monotonically improving policies and value functions:
$$
\pi_{0}\rightarrow V_{\pi_{0}}\rightarrow\pi_{1}\rightarrow V_{\pi_{1}}\rightarrow\dots\rightarrow\pi_{*}\rightarrow V_{\pi_{*}},
$$
where $V_{\pi_{0}}(s)\leq V_{\pi_{1}}(s)\leq \cdots\leq V_{\pi_{*}}(s)$ for all $s\in\mathcal{S}$.
\subsection{Relative Entropy Policy Search and Maximum a Posterior Policy Optimization}
The REPS (Relative Entropy Policy Search) algorithm \cite{10.5555/2898607.2898863} can be obtained as a special case of the update rule by replacing $\pi_{\mathrm{old}}$ with the observed data distribution and the coefficient $C_{\pi_{\mathrm{old}}}$ with the Lagrange multiplier $\eta$. However, the REPS algorithm is based on finite MDPs with discrete actions and not extendable to general continuous policies. A similar closed-form update rule has also been derived in the MPO (Maximum a posterior Policy Optimization) algorithm \cite{abdolmaleki2018maximum} in its E-step for evaluating a variational policy, which is then used to optimize policy parameters.

Our policy update rule is different from the previous work because it directly expresses the new policy as a closed-form function of the current policy. That means the policy update can be accurately calculated using the current policy without involving policy gradient or policy optimization. Especially, the proposed update rule provides an explicit formula for determining the coefficient $C_{\pi_{\mathrm{old}}}$ and guarantees monotonic improvement on performance. However, the update rules in \cite{10.5555/2898607.2898863, abdolmaleki2018maximum} need to numerically determine the optimal Lagrangian multiplier $\eta$, which requires a costly nonlinear optimization in the inner loop and no monotonic improvement is guaranteed.
\subsection{Soft Actor-Critic}
The SAC (Soft Actor-Critic) algorithm \cite{pmlr-v80-haarnoja18b,sac-arxiv} can also be derived as a special case of the policy update rule. Note that the update rule (\ref{update_formula}) can be expressed as a Gibbs measure (Boltzmann distribution in case of discrete actions):
\begin{equation}
\label{gibbs_explanation}
\begin{aligned}
	\pi_{\mathrm{new}}(a|s)=&\ \pi_{\mathrm{old}}(a|s)\frac{e^{A_{\pi_{\mathrm{old}}}(s,a)/C_{\pi_{\mathrm{old}}}}}{\mathbb{E}_{a\sim\pi_{\mathrm{old}}}\left[e^{A_{\pi_{\mathrm{old}}}(s,a)}/C_{\pi_{\mathrm{old}}}\right]}\\
	=&\ \pi_{\mathrm{old}}(a|s)\frac{e^{Q_{\pi_{\mathrm{old}}}(s,a)/C_{\pi_{\mathrm{old}}}}}{\mathbb{E}_{a\sim\pi_{\mathrm{old}}}\left[e^{Q_{\pi_{\mathrm{old}}}(s,a)/C_{\pi_{\mathrm{old}}}}\right]} \\
	=&\ \frac{1}{Z}\exp\big\{\frac{Q_{\pi_{\mathrm{old}}}(s,a)}{C_{\pi_{\mathrm{old}}}}+\log\pi_{\mathrm{old}}(a|s)\big\},
\end{aligned}
\end{equation}
where $Z=\mathbb{E}_{a\sim\pi_{\mathrm{old}}}\left[e^{Q_{\pi_{\mathrm{old}}}(s,a)/C_{\pi_{\mathrm{old}}}}\right]$ is the partition function.

To optimize a policy $\pi$, we can minimize the KL-Divergence between $\pi$ and $\pi_{\mathrm{new}}$:
\begin{equation}
\label{sac}
	\min_{\pi}\ D_{\mathrm{KL}}\Bigg(\pi(\cdot|s)\Bigg\Vert\frac{\exp\big(\frac{1}{C_{\pi_{\mathrm{old}}}}\widetilde{Q}_{\pi_{\mathrm{old}}}(s,\cdot)\big)}{Z}\Bigg),
\end{equation}
where $\widetilde{Q}_{\pi_{\mathrm{old}}}$ is the soft Q-function:
\begin{equation}
	\widetilde{Q}_{\pi_{\mathrm{old}}}(s,a)=Q_{\pi_{\mathrm{old}}}(s,a)+C_{\pi_{\mathrm{old}}}\log\pi_{\mathrm{old}}(a|s).
\end{equation}
Replacing $C_{\pi_{\mathrm{old}}}$ with a temperature parameter $\alpha$, we immediately get the SAC algorithm  \cite{sac-arxiv}.

A slight difference of the algorithm \eqref{sac} than SAC is that it minimizes the policy entropy instead of maximizing it. Note that the soft state value function derived from our update rule is given by
\begin{equation}
	\widetilde{V}_{\pi}(s)=\mathbb{E}_{a\sim\pi}[\widetilde{Q}_{\pi}(s,a)]=V_{\pi}(s)-C_{\pi}\mathcal{H}(\pi(\cdot|s)),
\end{equation}
where $\mathcal{H}(\pi(\cdot|s))$ is the policy entropy. Since $C_{\pi}$ is always positive, the policy entropy is penalized in the soft state value function. Thus, applying \eqref{sac} will minimize the policy entropy. This is reasonable because the policy distribution should be concentrating more and more on the optimal action as the policy improves monotonically.

The derivation of SAC also verifies that the update rule is essentially off-policy.
\section{Limitations and Discussions}
\subsection{Tightness of the Bound in Terms of $\gamma$}
The bound in Theorem \ref{new_bound} improves prior analysis in terms of KL-Divergence, but not in terms of $\gamma$, which could be a limitation of the policy update rule. Compared to the bound in TRPO, the improvement is at a cost of $\gamma/(4(1-\gamma)^2)$. When $\gamma$ is close to 1, the penalty coefficient $C_{\pi_{\mathrm{old}}}$ for the KL-Divergence can be large, resulting in small step sizes for policy updates. While $C_{\pi_{\mathrm{old}}}$ can be tuned to allow larger step-sizes in practice, a proven bound that is tighter in terms of $\gamma$ will be an interesting direction for future work.
\subsection{Monotonic Guarantee and Function Approximation}
The policy update rule is a closed-form solution, so it assumes an exact advantage function and an exact maximum of its absolute value. In large MDPs, these quantities generally need to be estimated by function approximators. The use of function approximatiors will inevitably introduce errors and can undermine the monotonic improvement guarantee. While our goal is to provide the theory, we would like to clarify this to encourage the development of efficient algorithms using function approximation. We also look forward to new RL theories building upon the update rule given its simplicity and wide connections with prior RL approaches.
\subsection{Simultaneous Update for Multi-Agent RL}
The extension of the update rule to multi-agent RL requires agents to take turns updating their policies. Thus, the learning process could be slow if there are many agents. From Equation \eqref{ma_update_rule} we see that the main reason for this requirement is that we need to make sure $\pi_{\mathrm{new}}^{-i}=\pi_{\mathrm{old}}^{-i}$. We believe that relaxing this requirement so as for the agents to update policies simultaneously without jeopardizing the monotonic improvement guarantee is worth studying in the future.
\section{Conclusion}
We have presented a closed-form update rule for general stochastic policy optimization with monotonic improvement guarantee. A new theoretical result has been provided by relating the lower bound of the performance to an expected KL-Divergence, which closes the gap between theory and practice in the literature. Based on the theoretical result, calculus of variation has been introduced to derive the policy update rule. Furthermore, we have proved that the policy update rule is extendable to cooperative multi-agent RL when agents take turns performing policy updates. Since the proposed update rule is analytical, we hope that it serves as a stepping stone for future work on novel RL theories and principled RL algorithms using parametric or non-parametric policies.

\section*{Acknowledgements}
This material is based upon work supported by the National Science Foundation under Grant No. ECCS 1917275.

\nocite{langley00}

\bibliography{ref}
\bibliographystyle{icml2022}

%%%%%%%%%%%%%%%%%%%%%%%%%%%%%%%%%%%%%%%%%%%%%%%%%%%%%%%%%%%%%%%%%%%%%%%%%%%%%%%
%%%%%%%%%%%%%%%%%%%%%%%%%%%%%%%%%%%%%%%%%%%%%%%%%%%%%%%%%%%%%%%%%%%%%%%%%%%%%%%
% APPENDIX
%%%%%%%%%%%%%%%%%%%%%%%%%%%%%%%%%%%%%%%%%%%%%%%%%%%%%%%%%%%%%%%%%%%%%%%%%%%%%%%
%%%%%%%%%%%%%%%%%%%%%%%%%%%%%%%%%%%%%%%%%%%%%%%%%%%%%%%%%%%%%%%%%%%%%%%%%%%%%%%
\newpage
\appendix
\onecolumn
\section{Proof of Policy Performance Bound}
\label{appendix-A}
This proof uses techniques from the proof of Lemma 3. in \cite{pmlr-v37-schulman15}, exploiting them to derive a new bound that relates to an average divergence between policies, $\pi', \pi$. An informal overview is as follows. First, using Lemma 1. in \cite{pmlr-v37-schulman15}, the gap between the surrogate and the objective is decomposed into the difference of two expected advantages over the policies $\pi', \pi$. Then, we use the \emph{coupling} technique to measure the coincidence of two trajectories resulted from $\pi', \pi$ before an arbitrary timestep $t$. Finally, we constrain the gap to an average KL-Divergence using Pinsker's inequality.
\\

\begin{definition}[Notations]
\label{notations}
We consider a Markov decision process with a continuous state space. The following definitions and notations will be used.

1. Probability density function (PDF) of the state at timestep $t$ given the policy $\pi$: $$\rho^{\pi}_{t}(s)=PDF(s_t=s|\pi).$$
Note that $\rho^{\pi}_{0}(s)=\rho_{0}(s)$ is the PDF of the initial state $s_0$, which is independent of $\pi$.

2. Discounted state visitation PDF: 
\begin{equation}
\begin{split}
	d^{\pi}(s)=&\ (1-\gamma)\big[\rho^{\pi}_{0}(s)+\gamma\rho^{\pi}_{1}(s)+\gamma^{2}\rho^{\pi}_{2}(s)+\cdots\big] \\
	=&\ (1-\gamma)\sum_{t=0}^{\infty}\gamma^{t}\rho^{\pi}_{t}(s).
\end{split}
\end{equation}

3. One-step state transition density given the policy $\pi$:
\begin{equation}
\label{one_step_trans_pdf}
	\nu_{\pi}(s'|s)=\int_{\mathcal{A}} p(s'|s,a)\pi(a|s)da.
\end{equation}

4. $t$-step state transition density given the policy $\pi$ (the Chapman Kolmogorov equation):
\begin{equation}
	\nu_{\pi}^{t}(s'|s)=\int_{\mathcal{S}}\nu_{\pi}^{m}(s'|\tilde{s})\nu_{\pi}^{t-m}(\tilde{s}|s)d\tilde{s},
\end{equation}
where $0\leq m\leq t$, and $\nu_{\pi}^{0}(s'|s)$ is a Dirac delta distribution:
\begin{equation}
	\nu_{\pi}^{0}(s'|s)=
\begin{cases}
	\infty,\ \text{if}\ s'=s,\\
	0,\ \text{otherwise}.
\end{cases}
\end{equation}
Note that $\nu_{\pi}^{0}(s'|s)$ is independent of the policy $\pi$, and thus $\nu_{\pi}^{0}(s'|s)=\nu_{\pi'}^{0}(s'|s)$.

5. Discounted state transition PDF given the policy $\pi$:
\begin{equation}
\label{mu_definition}
\begin{split}
	\mu_{\pi}(s'|s)=&(1-\gamma)\Big[\nu_{\pi}^{0}(s'|s)+\gamma\nu_{\pi}^{1}(s'|s)+\gamma^{2}\nu_{\pi}^{2}(s'|s)+\cdots\Big] \\
	=&(1-\gamma)\sum_{t=0}^{\infty}\gamma^{t}\nu_{\pi}^{t}(s'|s).
\end{split}
\end{equation}
Then, the discounted visitation PDF can be written as
\begin{equation}
	d^{\pi}(s')=\int_{\mathcal{S}}\rho_{0}(s)\mu_{\pi}(s'|s)ds.
\end{equation}

6. Surrogate model:
$$L_{\pi}(\pi')=J(\pi)+\frac{1}{1-\gamma}\mathbb{E}_{s\sim d^{\pi}, a\sim\pi'}[A_{\pi}(s,a)]$$

%7. Joint policy $\pi(a|s)$ and others' joint policy $\pi^{-i}(a^{-i}|s)$:
%$$\pi(a|s)=\prod_{i\in\mathcal{N}}\int_{o^{i}}\pi^{i}(a^{i}|o^{i})P_{o}^{i}(o^{i}|s)do^{i}$$
%$$\pi^{-i}(a^{-i}|s)=\prod_{j\in\mathcal{N}\setminus\{i\}}\int_{o^{j}}\pi^{j}(a^{j}|o^{j})P_{o}^{j}(o^{j}|s)do^{j}$$

7. Function spaces: For an open set $U\subset\mathbb{R}^{d}$, we define 
\begin{equation}
	C^{k}(U):=\{\text{Functions}\ u: U\rightarrow\mathbb{R}\text{ that are $k$-times continuously differentiable function on }U\}.
\end{equation}
%We also define the space of infinitely differentiable functions as follows:
%$$C^{\infty}(U) := \bigcap_{k=1}^{\infty} C^{k}(U).$$
\end{definition}

We start by introducing the definition of $\alpha$-\emph{coupled policies} from the Definition 1 in \cite{pmlr-v37-schulman15} with some changes.

\begin{definition}[$\alpha$-coupled policies]
\label{coupling}
A coupling of two probability distributions $\mu$ and $\nu$ is a pair of random variables
($X,Y)$ defined on a single probability space such that the marginal distribution
of $X$ is $\mu$ and the marginal distribution of $Y$ is $\nu$ \cite{LevinPeresWilmer2006}.

The policies $\pi'(a'|s)$ and $\pi(a|s)$ are called $\alpha$-coupled if they define a coupling of $(\pi',\pi)$ such that 
\begin{equation}
	P(a'\neq a|s)\leq \alpha(s).
\end{equation}
Numerically, $\alpha$-coupling means that the actions $a'$ and $a$ given state $s$ match with probability of at least $1-\alpha(s)$ when their samples are drawn using the same seed.

The technique of coupling is useful because it relates two policies to their total variation distance. According to the lemma 4.7 in \cite{LevinPeresWilmer2006}, for policies $\pi'$ and $\pi$, there exists a coupling that satisfies
\begin{equation}
\label{TV}
	D_{\mathrm{TV}}[\pi'||\pi](s) = \inf\{P(a'\neq a|s), \text{$a'$ and $a$ is a coupling of $\pi'$ and $\pi$}\}.
\end{equation}
where $D_{\mathrm{TV}}[\pi'||\pi](s)$ represents the total variation distance between policies $\pi'$ and $\pi$ given the state $s$. This means that $D_{\mathrm{TV}}[\pi'||\pi](s)$ is the infimum of the probability $P(a'\neq a|s)$, and therefore we can select $\alpha(s)$ to be $D_{\mathrm{TV}}[\pi'||\pi](s)$.

Note that our definition of $\alpha(s)$, depending on the state $s$, is different from the definition in \cite{pmlr-v37-schulman15}, which is the maximum over the state space, i.e. $\max_{s\in\mathcal{S}}\alpha(s)$.
\end{definition}

Next, we present a lemma from \cite{Kakade02approximatelyoptimal} and \cite{pmlr-v37-schulman15} that shows that the performance difference between two arbitrary policies can be expressed as an expected advantage of one policy over a trajectory resulted from the other.

\begin{lemma}
\label{lemma1}
Given two policies $\pi',\pi$, we have
\begin{equation}
	J(\pi')=J(\pi)+\frac{1}{1-\gamma}\mathbb{E}_{s\sim d^{\pi'},a\sim\pi'}\left[A_{\pi}(s,a)\right].
\end{equation}
\end{lemma}
\begin{proof}
Note that $A_{\pi}(s,a)=\mathbb{E}_{s'\sim P(\cdot|s,a)}\left[r(s,a)+\gamma V_{\pi}(s')-V_{\pi}(s)\right]$. Therefore,
\begin{equation}
\begin{split}
	&\ \mathbb{E}_{s\sim d^{\pi'},a\sim\pi'}\left[A_{\pi}(s,a)\right] \\
	=&\ \mathbb{E}_{s\sim d^{\pi'},a\sim\pi',s'\sim P}\left[r(s,a)+\gamma V_{\pi}(s')-V_{\pi}(s)\right] \\
	=&\ (1-\gamma)\mathbb{E}_{s_{t}\sim \rho_{t}^{\pi'}, a_{t}\sim\pi', s_{t+1}\sim P}\left[\sum_{t=0}^{\infty}\gamma^{t}\Big(r(s_{t},a_{t})+\gamma V_{\pi}(s_{t+1})-V_{\pi}(s_{t})\Big)\right] \\
	=&\ (1-\gamma)\mathbb{E}_{s_{t}\sim \rho_{t}^{\pi'}, a_{t}\sim\pi'}\left[-V_{\pi}(s_{0})+\sum_{t=0}^{\infty}\gamma^{t}r(s_{t},a_{t})\right] \\
	=&\ (1-\gamma)\left(-\mathbb{E}_{s_{0}\sim\rho_0}\left[V_{\pi}(s_{0})\right]+\mathbb{E}_{s_{t}\sim \rho_{t}^{\pi'}, a_{t}\sim\pi'}\left[\sum_{t=0}^{\infty}\gamma^{t}r(s_{t},a_{t})\right]\right) \\
	=&\ (1-\gamma)\left[-J(\pi)+J(\pi')\right]
\end{split}
\end{equation}
Rearranging it, the result follows.
\end{proof}

\begin{lemma}
\label{lemma2}
Given two stochastic policies $\pi',\pi$ and their discounted state transition PDFs, $\mu_{\pi'}(s'|s),\mu_{\pi}(s'|s)$, the following inequality holds:
\begin{equation}
\begin{split}
	\int_{\mathcal{S}}\big\vert\mu_{\pi'}(s'|s)-\mu_{\pi}(s'|s)\big\vert ds'\leq \frac{2\gamma^{2}}{1-\gamma}\int_{\mathcal{S}}\mu_{\pi}(s'|s)D_{\mathrm{TV}}[\pi'||\pi](s')ds'.
\end{split}
\end{equation}
\end{lemma}
\begin{proof}
First note that
\begin{equation}
\begin{split}
	&\ \gamma\int_{\mathcal{S}}\nu_{\pi}(s'|\bar{s})\mu_{\pi}(\bar{s}|s)d\bar{s} \\
	=&\ \gamma\int_{\mathcal{S}}\nu_{\pi}(s'|\bar{s})(1-\gamma)\Big[\nu_{\pi}^{0}(\bar{s}|s)+\gamma\nu_{\pi}^{1}(\bar{s}|s)+\gamma^{2}\nu_{\pi}^{2}(\bar{s}|s)+\cdots\Big]d\bar{s} \\
	=&\ (1-\gamma)\Big[\gamma\nu_{\pi}^{1}(s'|s)+\gamma^{2}\nu_{\pi}^{2}(s'|s)+\gamma^{3}\nu_{\pi}^{3}(s'|s)+\cdots\Big] \\
	=&\ \mu_{\pi}(s'|s)-(1-\gamma)\nu_{\pi}^{0}(s'|s).
\end{split}
\end{equation}
Then, we have
\begin{equation}
\begin{aligned}
	&\ \gamma\iint_{\mathcal{S}\times\mathcal{S}}\mu_{\pi'}(s'|\tilde{s})\left[\nu_{\pi'}(\tilde{s}|\bar{s})-\nu_{\pi}(\tilde{s}|\bar{s})\right]\mu_{\pi}(\bar{s}|s)d\tilde{s}d\bar{s} \\
	=&\ \int_{\mathcal{S}}\left(\gamma\int_{\mathcal{S}}\mu_{\pi'}(s'|\tilde{s})\nu_{\pi'}(\tilde{s}|\bar{s})d\tilde{s}\right)\mu_{\pi}(\bar{s}|s)d\bar{s}-\int_{\mathcal{S}}\mu_{\pi'}(s'|\tilde{s})\left(\gamma\int_{\mathcal{S}}\nu_{\pi}(\tilde{s}|\bar{s})\mu_{\pi}(\bar{s}|s)d\bar{s}\right)d\tilde{s} \\
	=&\ \int_{\mathcal{S}}\left[\mu_{\pi'}(s'|\bar{s})-(1-\gamma)\nu_{\pi'}^{0}(s'|\bar{s})\right]\mu_{\pi}(\bar{s}|s)d\bar{s}-\int_{\mathcal{S}}\mu_{\pi'}(s'|\tilde{s})\left[\mu_{\pi}(\tilde{s}|s)-(1-\gamma)\nu_{\pi}^{0}(\tilde{s}|s)\right]d\tilde{s} \\
	=&\ {\color{blue}\cancel{\int_{\mathcal{S}}\mu_{\pi'}(s'|\bar{s})\mu_{\pi}(\bar{s}|s)d\bar{s}}} - (1-\gamma)\int_{\mathcal{S}}\nu_{\pi'}^{0}(s'|\bar{s})\mu_{\pi}(\bar{s}|s)d\bar{s}{\color{blue}\cancel{-\int_{\mathcal{S}}\mu_{\pi'}(s'|\tilde{s})\mu_{\pi}(\tilde{s}|s)d\tilde{s}}} + (1-\gamma)\int_{\mathcal{S}}\mu_{\pi'}(s'|\tilde{s})\nu_{\pi}^{0}(\tilde{s}|s)d\tilde{s} \\
	=&\ (1-\gamma)\Bigg[\int_{\mathcal{S}}\mu_{\pi'}(s'|\tilde{s})\nu_{\pi}^{0}(\tilde{s}|s)d\tilde{s}-\int_{\mathcal{S}}\nu_{\pi'}^{0}(s'|\bar{s})\mu_{\pi}(\bar{s}|s)d\bar{s}\Bigg]\ \ \ ({\color{blue}\text{Note that $\nu_{\pi}^{0}(\tilde{s}|s)=\nu_{\pi'}^{0}(\tilde{s}|s)$}}) \\
	=&\ \frac{1-\gamma}{\gamma}\Bigg[\Big(\mu_{\pi'}(s'|s)-(1-\gamma)\nu_{\pi'}^{0}(s'|s)\Big)-\Big(\mu_{\pi}(s'|s)-(1-\gamma)\nu_{\pi}^{0}(s'|s)\Big)\Bigg] \\
	=&\ \frac{1-\gamma}{\gamma}\left[\mu_{\pi'}(s'|s)-\mu_{\pi}(s'|s)\right].
\end{aligned}
\end{equation}
Rearranging the equation, we have
\begin{equation}
	\mu_{\pi'}(s'|s)-\mu_{\pi}(s'|s)=\frac{\gamma^{2}}{1-\gamma}\iint_{\mathcal{S}\times\mathcal{S}}\mu_{\pi'}(s'|\tilde{s})\left[\nu_{\pi'}(\tilde{s}|\bar{s})-\nu_{\pi}(\tilde{s}|\bar{s})\right]\mu_{\pi}(\bar{s}|s)d\tilde{s}d\bar{s}.
\end{equation}
Recalling the definition of one-step state transition density in Equation \eqref{one_step_trans_pdf}, we have
\begin{equation}
\begin{split}
	\nu_{\pi'}(\tilde{s}|\bar{s})-\nu_{\pi}(\tilde{s}|\bar{s})=\int_{\mathcal{A}}p(\tilde{s}|\bar{s},\bar{a})\left[\pi'(\bar{a}|\bar{s})-\pi(\bar{a}|\bar{s})\right]d\bar{a}.
\end{split}
\end{equation}
Then, we have
\begin{equation}
\begin{split}
	&\ \int_{\mathcal{S}}\vert\mu_{\pi'}(s'|s)-\mu_{\pi}(s'|s)\vert ds' \\
	\leq&\ \frac{\gamma^{2}}{1-\gamma}\iiiint_{\mathcal{S}\times\mathcal{S}\times\mathcal{S}\times\mathcal{A}}\mu_{\pi'}(s'|\tilde{s})p(\tilde{s}|\bar{s},\bar{a})\vert\pi'(\bar{a}|\bar{s})-\pi(\bar{a}|\bar{s})\vert\mu_{\pi}(\bar{s}|s)ds' d\tilde{s}d\bar{s}d\bar{a} \\
	=&\ \frac{\gamma^{2}}{1-\gamma}\int_{\mathcal{S}}\mu_{\pi'}(s'|\tilde{s})ds'\iiint_{\mathcal{S}\times\mathcal{S}\times\mathcal{A}}p(\tilde{s}|\bar{s},\bar{a})\vert\pi'(\bar{a}|\bar{s})-\pi(\bar{a}|\bar{s})\vert\mu_{\pi}(\bar{s}|s)d\tilde{s}d\bar{s}d\bar{a} \\
	=&\ \frac{\gamma^{2}}{1-\gamma}\int_{\mathcal{S}}p(\tilde{s}|\bar{s},\bar{a})d\tilde{s}\iint_{\mathcal{S}\times\mathcal{A}}\vert\pi'(\bar{a}|\bar{s})-\pi(\bar{a}|\bar{s})\vert\mu_{\pi}(\bar{s}|s)d\bar{s}d\bar{a} \\
	=&\ \frac{\gamma^{2}}{1-\gamma}\int_{\mathcal{S}}\mu_{\pi}(\bar{s}|s)\int_{\mathcal{A}}\vert\pi'(\bar{a}|\bar{s})-\pi(\bar{a}|\bar{s})\vert d\bar{a}d\bar{s} \\
	=&\ \frac{2\gamma^{2}}{1-\gamma}\int_{\mathcal{S}}\mu_{\pi}(\bar{s}|s)D_{\mathrm{TV}}[\pi'||\pi](\bar{s})d\bar{s}.
\end{split}
\end{equation}
Replacing all $\bar{s}$ with $s'$, the result follows.
\end{proof}

\begin{lemma}
\label{lemma3}
Let $\overline{\alpha}_{t}$ and $\gamma$ be any real numbers within $\overline{\alpha}_{t}\in [0,1], \forall t\in\mathbb{N}$ and $\gamma\in[0.5,1)$ , the following inequality holds: 
\begin{equation}
\label{TV2KL}
	(1-\gamma)^{2}\sum_{t=0}^{\infty}\gamma^{t}\overline{\alpha}_{t}\overline{\alpha}_{0t} \leq \sum_{t=0}^{\infty}\gamma^{t}\overline{\alpha}_{t}^{2}.
\end{equation}
where $\overline{\alpha}_{0t}=1-\prod_{i=0}^{t}(1-\overline{\alpha}_{i})$.

\begin{proof}
First note that $\overline{\alpha}_{0t}$ can be expressed as
\begin{equation}
\label{1_minus_prod}
	\overline{\alpha}_{0t}=\overline{\alpha}_{t}+(1-\overline{\alpha}_{t})\overline{\alpha}_{t-1}+(1-\overline{\alpha}_{t})(1-\overline{\alpha}_{t-1})\overline{\alpha}_{t-2}+\cdots+\prod_{i=1}^{t}(1-\overline{\alpha}_{i})\overline{\alpha}_{0},
\end{equation}
or in a recursive form:
\begin{equation}
\label{1_minus_prod_sim}
	\overline{\alpha}_{0t}=\overline{\alpha}_{t}+(1-\overline{\alpha}_{t})\overline{\alpha}_{0t-1},
\end{equation}
where $\overline{\alpha}_{00}=\overline{\alpha}_{0}$.
Then, we have
\begin{equation}
\label{extract_KL}
\begin{split}
	&\ (1-\gamma)^{2}\sum_{t=0}^{\infty}\gamma^{t}\overline{\alpha}_{t}\overline{\alpha}_{0t} \\
	=&\ (1-\gamma)^{2}\sum_{t=0}^{\infty}\gamma^{t}\overline{\alpha}_{t}^{2}+(1-\gamma)^{2}\sum_{t=1}^{\infty}\gamma^{t}\overline{\alpha}_{t}(1-\overline{\alpha}_{t})\overline{\alpha}_{0t-1} \\
	=&\ \sum_{t=0}^{\infty}\gamma^{t}\overline{\alpha}_{t}^{2}-(2\gamma-\gamma^{2})\sum_{t=0}^{\infty}\gamma^{t}\overline{\alpha}_{t}^{2}+(1-\gamma)^{2}\sum_{t=1}^{\infty}\gamma^{t}\overline{\alpha}_{t}(1-\overline{\alpha}_{t})\overline{\alpha}_{0t-1} \\
	=& \sum_{t=0}^{\infty}\gamma^{t}\overline{\alpha}_{t}^{2}-\left[(2\gamma-\gamma^{2})\sum_{t=0}^{\infty}\gamma^{t}\overline{\alpha}_{t}^{2}-(1-\gamma)^{2}\sum_{t=1}^{\infty}\gamma^{t}\overline{\alpha}_{t}(1-\overline{\alpha}_{t})\overline{\alpha}_{0t-1}\right]
\end{split}
\end{equation}

For the inequality \eqref{TV2KL} to hold, we only need to prove that the subtrahend on the rightest-hand side of \eqref{extract_KL} is greater than 0. Note that
\begin{equation}
\label{major_inequality}
\begin{split}
	&\ (2\gamma-\gamma^{2})\sum_{t=0}^{\infty}\gamma^{t}\overline{\alpha}_{t}^{2}-(1-\gamma)^{2}\sum_{t=1}^{\infty}\gamma^{t}\overline{\alpha}_{t}(1-\overline{\alpha}_{t})\overline{\alpha}_{0t-1} \\
	=&\ \sum_{t=0}^{\infty}\gamma^{t}\Big[\gamma+\gamma(1-\gamma)\Big]\overline{\alpha}_{t}^{2}-(1-\gamma)^{2}\sum_{t=1}^{\infty}\gamma^{t}\overline{\alpha}_{t}(1-\overline{\alpha}_{t})\overline{\alpha}_{0t-1} \\
	=&\ \sum_{t=0}^{\infty}\gamma^{t}\left[(1-\gamma)\sum_{i=1}^{\infty}\gamma^{i}+(1-\gamma)^2\sum_{i=1}^{\infty}\gamma^{i}\right]\overline{\alpha}_{t}^{2}-(1-\gamma)^{2}\sum_{t=1}^{\infty}\gamma^{t}\overline{\alpha}_{t}(1-\overline{\alpha}_{t})\overline{\alpha}_{0t-1} \\
	=&\ (1-\gamma)^{2}\sum_{t=0}^{\infty}\gamma^{t}\sum_{i=1}^{\infty}\gamma^{i}\left[\frac{1}{1-\gamma}+1\right]\overline{\alpha}_{t}^{2}-(1-\gamma)^{2}\sum_{t=1}^{\infty}\gamma^{t}\overline{\alpha}_{t}(1-\overline{\alpha}_{t})\overline{\alpha}_{0t-1} \ \ \ {\color{blue}\Big(\frac{1}{1-\gamma}=1+\gamma+\gamma^2+\cdots\Big)} \\
	=&\ \gamma(1-\gamma)\overline{\alpha}_{0}^{2} + (1-\gamma)^{2}\sum_{t=1}^{\infty}\gamma^{t}\left[\sum_{n=0}^{t-1}(t-n)\overline{\alpha}_{n}^{2}+\frac{\gamma}{1-\gamma}\overline{\alpha}_{t}^{2}-\overline{\alpha}_{t}(1-\overline{\alpha}_{t})\overline{\alpha}_{0t-1}\right] \\
	\geq&\ \gamma(1-\gamma)^2\overline{\alpha}_{0}^{2} + (1-\gamma)^{2}\sum_{t=1}^{\infty}\gamma^{t}\left[\sum_{n=0}^{t-1}(t-n)\overline{\alpha}_{n}^{2}+\overline{\alpha}_{t}^{2}-\overline{\alpha}_{t}\overline{\alpha}_{0t-1}\right].\ \ \ (\text{since}\ \gamma>1-\gamma\ \text{when}\ \gamma\geq 0.5)\\
\end{split}
\end{equation}
In the expanded form, the rightest-hand side of \eqref{major_inequality} can be expressed as
\begin{equation}
\label{expanded_form}
\begin{split}
	&\ \gamma(1-\gamma)^2\overline{\alpha}_{0}^{2} + (1-\gamma)^{2}\sum_{t=1}^{\infty}\gamma^{t}\left[\sum_{n=0}^{t-1}(t-n)\overline{\alpha}_{n}^{2}+\overline{\alpha}_{t}^{2}-\overline{\alpha}_{t}\overline{\alpha}_{0t-1}\right] \\
	=&\ \gamma(1-\gamma)^{2}\Big[\overline{\alpha}_{0}^{2}+\overline{\alpha}_{0}^{2}+\overline{\alpha}_{1}^{2}-\overline{\alpha}_{1}\overline{\alpha}_{0}\Big] + \\
	&\ \gamma^{2}(1-\gamma)^{2}\Big[2\overline{\alpha}_{0}^{2}+\overline{\alpha}_{1}^{2}+\overline{\alpha}_{2}^{2}-\overline{\alpha}_{2}\overline{\alpha}_{01}\Big] + \\
	&\ \gamma^{3}(1-\gamma)^{2}\Big[3\overline{\alpha}_{0}^{2}+2\overline{\alpha}_{1}^{2}+\overline{\alpha}_{2}^{2}+\overline{\alpha}_{3}^{2}-\overline{\alpha}_{3}\overline{\alpha}_{02}\Big] + \\
	&\ \gamma^{4}(1-\gamma)^{2}\Big[4\overline{\alpha}_{0}^{2}+3\overline{\alpha}_{1}^{2}+2\overline{\alpha}_{2}^{2}+\overline{\alpha}_{3}^{2}+\overline{\alpha}_{4}^{2}-\overline{\alpha}_{4}\overline{\alpha}_{03}\Big] + \\
	&\ \cdots 
\end{split}
\end{equation}
Since
\begin{equation}
	\gamma^{t}(1-\gamma)^2 = \gamma^{t}(1-\gamma)^2(1-\gamma+\gamma)=\gamma^{t}(1-\gamma)^3+\gamma^{t+1}(1-\gamma)^2,
\end{equation}
Equation \eqref{expanded_form} can be rewritten as
\begin{equation}
\begin{split}
	&\ \gamma(1-\gamma)^{\color{red}3}\Big[2\overline{\alpha}_{0}^{2}+\overline{\alpha}_{1}^{2}-\overline{\alpha}_{1}\overline{\alpha}_{0}\Big] + \\
	&\ \gamma^{2}(1-\gamma)^{3}\Big[4\overline{\alpha}_{0}^{2}+2\overline{\alpha}_{1}^{2}+\overline{\alpha}_{2}^{2}-\overline{\alpha}_{1}\overline{\alpha}_{0}-\overline{\alpha}_{2}\overline{\alpha}_{01}\Big] + \\
	&\ \gamma^{3}(1-\gamma)^{3}\Big[7\overline{\alpha}_{0}^{2}+4\overline{\alpha}_{1}^{2}+2\overline{\alpha}_{2}^{2}+\overline{\alpha}_{3}^{2}-\overline{\alpha}_{1}\overline{\alpha}_{0}-\overline{\alpha}_{2}\overline{\alpha}_{01}-\overline{\alpha}_{3}\overline{\alpha}_{02}\Big] + \\
	&\ \gamma^{4}(1-\gamma)^{3}\Big[11\overline{\alpha}_{0}^{2}+7\overline{\alpha}_{1}^{2}+4\overline{\alpha}_{2}^{2}+2\overline{\alpha}_{3}^{2}+\overline{\alpha}_{4}^{2}-\overline{\alpha}_{1}\overline{\alpha}_{0}-\overline{\alpha}_{2}\overline{\alpha}_{01}-\overline{\alpha}_{3}\overline{\alpha}_{02}-\overline{\alpha}_{4}\overline{\alpha}_{03}\Big] + \\
	&\ \cdots \\
	=&\ (1-\gamma)^{3}\sum_{t=1}^{\infty}\gamma^{t}\Bigg[a_{t}\overline{\alpha}_{0}^{2}+\sum_{i=1}^{t}\Big(a_{t-i}\overline{\alpha}_{i}^{2}-\overline{\alpha}_{i}\overline{\alpha}_{0i-1}\Big)\Bigg] \\
	=&\ (1-\gamma)^{3}\sum_{t=1}^{\infty}\gamma^{t}H_{t}
\end{split}
\end{equation}
where
\begin{equation}
\label{a_i}
	a_{t}=1+\sum_{j=0}^{t}j,
\end{equation}
and
\begin{equation}
\label{H_function}
	H_{t}=a_{t}\overline{\alpha}_{0}^{2}+\sum_{i=1}^{t}\Big(a_{t-i}\overline{\alpha}_{i}^{2}-\overline{\alpha}_{i}\overline{\alpha}_{0i-1}\Big),
\end{equation}
Next, we prove $H_{t}\geq 0$ for all $t\in\mathbb{N}^{+}$ by using convex optimization. Decompose $H_{t}$ into:
\begin{equation}
\begin{split}
	H_{t}=&\ a_{t}\overline{\alpha}_{0}^{2}+\sum_{i=1}^{t-1}\Big(a_{t-i}\overline{\alpha}_{i}^{2}-\overline{\alpha}_{i}\overline{\alpha}_{0i-1}\Big)+a_{0}{\color{blue}\overline{\alpha}_{t}}^{2}-{\color{blue}\overline{\alpha}_{t}}\overline{\alpha}_{0t-1} \\
	=&\ h_{t-1}+a_{0}{\color{blue}\overline{\alpha}_{t}}^{2}-{\color{blue}\overline{\alpha}_{t}}\overline{\alpha}_{0t-1}
\end{split}
\end{equation}
Taking the partial derivative of $H_{t}$ with respect to $\overline{\alpha}_{t}$ and setting it to be zero, $H_{t}$ attains its minimum value, i.e.,
\begin{equation}
	H_{t}\geq h_{t-1}-\frac{1}{4a_{0}}\overline{\alpha}_{0t-1}^{2}.
\end{equation}
Denoting $b_{1}=1/4a_{0}$ and decomposing $h_{t-1}$, we get
\begin{equation}
\begin{split}
	H_{t}\geq&\ h_{t-2}+a_{1}\overline{\alpha}_{t-1}^{2}-\overline{\alpha}_{t-1}\overline{\alpha}_{0t-2}-b_{1}\overline{\alpha}_{0t-1}^{2} \\
	\geq&\ h_{t-2}+a_{1}\overline{\alpha}_{t-1}^{2}-\overline{\alpha}_{t-1}\overline{\alpha}_{0t-2}-b_{1}[\overline{\alpha}_{t-1}+\overline{\alpha}_{0t-2}]^{2} \\
	=&\ h_{t-2}+(a_{1}-b_{1})\overline{\alpha}_{t-1}^{2}-(2b_{1}+1)\overline{\alpha}_{t-1}\overline{\alpha}_{0t-2}-b_{1}\overline{\alpha}_{0t-2}^{2}
\end{split}
\end{equation}
Again, taking the partial derivative with respect to $\overline{\alpha}_{t-1}$ and setting it to be zero, we get
\begin{equation}
\begin{split}
	H_{t}\geq h_{t-2}-\frac{b_{1}+a_{1}b_{1}+1/4}{a_{1}-b_{1}}\overline{\alpha}_{0t-2}^{2}.
\end{split}
\end{equation}
Recursively, as long as $b_{i}\geq 0$ and $a_{i}-b_{i}\geq 0$ hold for all $i\leq t\in\mathbb{N}^{+}$, we can repeatedly apply the previous procedure and get
\begin{equation}
\label{H_t_inequality}
	H_{t}\geq h_{t-1}-b_{1}\overline{\alpha}_{0t-1}^{2}\geq h_{t-2}-b_{2}\overline{\alpha}_{0t-2}^{2}\geq \cdots \geq a_{t}\overline{\alpha}_{0}^{2}-b_{t}\overline{\alpha}_{0}^{2}\geq 0,
\end{equation}
where
\begin{equation}
	b_{i+1}=\frac{b_{i}+a_{i}b_{i}+1/4}{a_{i}-b_{i}},\ i=0,\dots,t-1,
\end{equation} 
$b_{0}=0$ and $a_{i}$ is defined in Equation \eqref{a_i}. Next, we prove $b_{i}\geq 0$ and $a_{i}-b_{i}\geq 0$ for all $i\leq t\in\mathbb{N}^{+}$. 

First, it is easy to manually verify that $b_{i}\geq 0$ and $a_{i}-b_{i}\geq 0$ when $i<15$. In addition, for $i=15$, we can verify that the following inequalities holds
\begin{equation}
\label{key_inequalities_lemma3}
	a_{i}-b_{i}\geq 4(b_{i}+\frac{1}{2})^{2},\ \ \ b_{i}\leq \frac{1}{4}i,
\end{equation}
since $a_{15}=121$ and $b_{15}\approx 3.6945$.

Next, we prove the inequalities in \eqref{key_inequalities_lemma3} holds for $i>15$. Note that
\begin{equation}
\label{b_diff}
\begin{aligned}
	b_{i+1}-b_{i}=&\ \frac{b_{i}+a_{i}b_{i}+1/4}{a_{i}-b_{i}}-b_{i}=\frac{(b_{i}+\frac{1}{2})^{2}}{a_{i}-b_{i}}, \\
	a_{i+1}-a_{i}=&\ i+1.
\end{aligned}
\end{equation}
Therefore, we have
\begin{equation}
\label{b_induction}
	b_{i+1}=b_{i}+\frac{(b_{i}+\frac{1}{2})^{2}}{a_{i}-b_{i}}\leq b_{i}+\frac{1}{4}\leq \frac{1}{4}(i+1),
\end{equation}
and
\begin{equation}
\label{ab_induction}
\begin{split}
	a_{i+1}-b_{i+1} =&\ i+1+a_{i}-b_{i}-\frac{(b_{i}+\frac{1}{2})^{2}}{a_{i}-b_{i}} \\
	\geq &\ i+1 + 4(b_{i}+\frac{1}{2})^{2} - \frac{1}{4} \\
	\geq &\ i+1 + 4\Big(b_{i+1}+\frac{1}{2}-\frac{1}{4}\Big)^{2} - \frac{1}{4} \\
	=&\ 4(b_{i+1}+\frac{1}{2})^{2} - 2b_{i+1}+i \\
	\geq &\ 4(b_{i+1}+\frac{1}{2})^{2} - \frac{1}{2}(i+1) + i\ \ \ (i>15) \\
	\geq &\ 4(b_{i+1}+\frac{1}{2})^{2}.
\end{split}
\end{equation}

Based on Equations \eqref{key_inequalities_lemma3}, \eqref{b_induction} and \eqref{ab_induction}, we can prove that $b_{i}\geq 0$ and $a_{i}-b_{i}\geq 0$ hold for $i\geq 15$ using mathematical induction. Combining the fact that they also hold for $i<15$, we have $b_{i}\geq 0$ and $a_{i}-b_{i}\geq 0$ for all $i\leq t\in\mathbb{N}^{+}$. As a result, the inequality \eqref{H_t_inequality} holds, i.e., $H_{t}\geq 0$, which concludes the proof.
\end{proof}
\end{lemma}

\begin{theorem}
\label{new_bound_appendix}
For any stochastic policies $\pi',\pi$ and discount factor $\gamma\in[0.5,1)$, the following bound holds:
\begin{equation}
\begin{split}
	&\vert J(\pi') - L_{\pi}(\pi')\vert \leq \frac{1}{1-\gamma}C_{\pi}\mathbb{E}_{s\sim d^{\pi}}\left[D_{\mathrm{KL}}[\pi'\Vert\pi](s)\right], \\
	&\text{where}\ \ C_{\pi}=\frac{\gamma^{2}\epsilon}{(1-\gamma)^{3}},\ \epsilon=\max_{s,a}|A_{\pi}(s,a)|.
\end{split}
\end{equation}
\end{theorem}
\begin{proof}[Proof]
Define $\overline{A}(s)$ to be the expected advantage of $\pi'$ over $\pi$ at state s:
\begin{equation}
\label{Abar}
	\overline{A}(s)=\mathbb{E}_{a\sim\pi'(\cdot|s)}\left[A_{\pi}(s,a)\right]
\end{equation}
Then, Lemma \ref{lemma1} can be rewritten as follows:
\begin{equation}
\label{A_bar}
	J(\pi')=J(\pi)+\frac{1}{1-\gamma}\mathbb{E}_{s\sim {\color{red}d^{\pi'}}}\left[\overline{A}(s)\right]=J(\pi)+\sum_{t=0}^{\infty}\gamma^{t}\mathbb{E}_{s_{t}\sim {\color{red}\rho_{t}^{\pi'}}}\left[\overline{A}(s_{t})\right].
\end{equation}
Note that the surrogate model can be written as
\begin{equation}
	L_{\pi}(\pi')=J(\pi)+\frac{1}{1-\gamma}\mathbb{E}_{s\sim {\color{red}d^{\pi}}}\left[\overline{A}(s)\right]=J(\pi)+\sum_{t=0}^{\infty}\gamma^{t}\mathbb{E}_{s_{t}\sim {\color{red}\rho_{t}^{\pi}}}\left[\overline{A}(s_{t})\right].
\end{equation}
Then, the difference between the surrogate $L_{\pi}(\pi')$ and the true objective $J(\pi')$ can be written as
\begin{equation}
\label{lemma_difference}
\begin{split}
	J(\pi') - L_{\pi}(\pi') & = \sum_{t=0}^{\infty}\gamma^{t}\Big[\mathbb{E}_{s_{t}\sim\rho_{t}^{\pi'}}\left[\overline{A}(s_{t})\right]-\mathbb{E}_{s_{t}\sim\rho_{t}^{\pi}}\left[\overline{A}(s_{t})\right]\Big] \\
	& = \sum_{{\color{red}t=1}}^{\infty}\gamma^{t}\Big[\mathbb{E}_{s_{t}\sim\rho_{t}^{\pi'}}\left[\overline{A}(s_{t})\right]-\mathbb{E}_{s_{t}\sim\rho_{t}^{\pi}}\left[\overline{A}(s_{t})\right]\Big].\ \ \ \text{(since $p_{0}^{\pi'}=p_{0}^{\pi}=\rho_{0}$)}
\end{split}
\end{equation}

Next, we split the proof into three parts. (1) By using the coupling technique, we decompose the difference terms in \eqref{lemma_difference}, i.e. $\mathbb{E}_{s_{t}\sim p_{t}^{\pi'}}\left[\overline{A}(s_{t})\right]-\mathbb{E}_{s_{t}\sim p_{t}^{\pi}}\left[\overline{A}(s_{t})\right]$, to derive an equivalent expression. (2) Based on the result from the first part, we use Lemma \ref{lemma2} to derive an upper bound of $\vert J(\pi') - L_{\pi}(\pi')\vert$, which depends on a bunch of state-dependent total variation distances of $\pi',\pi$. (3) We relate the bound derived from the second part to the expected KL-Divergence between $\pi',\pi$.

i) The first part of the proof is given as follows.

We will use techniques from the proof of Lemma 3. in \cite{pmlr-v37-schulman15} to measure the coincidence of two trajectories resulted from $\pi', \pi$ before an arbitrary timestep $t$.
Let $n_t$ denote the number of times that $a'_{i}\neq a_{i}|s_{i}$ at state $s_i$ for $i<t$. For instance, $n_t=0$ means the trajectories $\tau,\tau'$ completely match before timestep $t$, i.e., $a'_{i}=a_{i}|s_{i}$ for all $i<t$.

The expected advantage at state $s_t$ on the trajectory $\tau'\sim\pi'$ decomposes as follows:
\begin{equation}
\label{exp_adv_tau_prime}
	\mathbb{E}_{s_{t}\sim\rho_{t}^{\pi'}}\left[\overline{A}(s_{t})\right]=P(n_{t}=0)\mathbb{E}_{s_{t}\sim\rho_{t}^{\pi'}|n_{t}=0}\left[\overline{A}(s_{t})\right] + P(n_{t}>0)\mathbb{E}_{s_{t}\sim\rho_{t}^{\pi'}|n_{t}>0}\left[\overline{A}(s_{t})\right].
\end{equation}
The expected advantage on the trajectory $\tau\sim\pi$ decomposes similarly:
\begin{equation}
\label{exp_adv_tau}
	\mathbb{E}_{s_{t}\sim\rho_{t}^{\pi}}\left[\overline{A}(s_{t})\right]=P(n_{t}=0)\mathbb{E}_{s_{t}\sim\rho_{t}^{\pi}|n_{t}=0}\left[\overline{A}(s_{t})\right] + P(n_{t}>0)\mathbb{E}_{s_{t}\sim\rho_{t}^{\pi}|n_{t}>0}\left[\overline{A}(s_{t})\right].
\end{equation}
Subtracting Equation \eqref{exp_adv_tau} from \eqref{exp_adv_tau_prime}, we get
\begin{equation}
\label{exp_adv_diff}
	\mathbb{E}_{s_{t}\sim\rho_{t}^{\pi'}}\left[\overline{A}(s_{t})\right] - \mathbb{E}_{s_{t}\sim\rho_{t}^{\pi}}\left[\overline{A}(s_{t})\right] = P(n_{t}>0)\Big(\mathbb{E}_{s_{t}\sim\rho_{t}^{\pi'}|n_{t}>0}\left[\overline{A}(s_{t})\right] - \mathbb{E}_{s_{t}\sim\rho_{t}^{\pi}|n_{t}>0}\left[\overline{A}(s_{t})\right]\Big),
\end{equation}
because $\mathbb{E}_{s_{t}\sim {\color{red}\rho_{t}^{\pi'}}|n_{t}=0}\left[\overline{A}(s_{t})\right]=\mathbb{E}_{s_{t}\sim {\color{red}\rho_{t}^{\pi}}|n_{t}=0}\left[\overline{A}(s_{t})\right]$ when $n_{t}=0$.

Note that
\begin{equation}
	n_{t}>0\Rightarrow
	\begin{cases}
		n_{t-1}=0\ \text{and}\ a'_{t-1}\neq a_{t-1}|s_{t-1}\ \text{for every $s_{t-1}$},\ \text{or} \\
		n_{t-1}>0,
	\end{cases}
\end{equation}
so we have
\begin{equation}
\begin{split}
	P(n_{t}>0) = &\ P(n_{t-1}=0)\cdot\mathbb{E}_{s_{t-1}\sim\rho_{t-1}^{\pi}}\left[P(a'_{t-1}\neq a_{t-1}|s_{t-1})\right] + P(n_{t-1}>0).
\end{split}
\end{equation}
In a recursive form, it can be expressed as:
\begin{equation}
\label{disagree_t}
\begin{split}
	P(n_{t}>0) = \sum_{i=0}^{t-1}P(n_{i}=0)\mathbb{E}_{s_{i}\sim\rho_{i}^{\pi}}\left[P(a'_{i}\neq a_{i}|s_{i})\right]
\end{split}
\end{equation}
Substituting \eqref{disagree_t} into \eqref{exp_adv_diff}, we get
\begin{equation}
\label{exp_adv_diff_upper_bound_almost}
\begin{split}
	&\ \mathbb{E}_{s_{t}\sim\rho_{t}^{\pi'}}\left[\overline{A}(s_{t})\right] - \mathbb{E}_{s_{t}\sim\rho_{t}^{\pi}}\left[\overline{A}(s_{t})\right] \\
	=&\ \sum_{i=0}^{t-1}P(n_{i}=0)\mathbb{E}_{s_{i}\sim\rho_{i}^{\pi}}\left[P(a'_{i}\neq a_{i}|s_{i})\right]\Big(\mathbb{E}_{s_{t}\sim\rho_{t}^{\pi'}|n_{i}=0}\left[\overline{A}(s_{t})\right] - \mathbb{E}_{s_{t}\sim\rho_{t}^{\pi}|n_{i}=0}\left[\overline{A}(s_{t})\right]\Big).
\end{split}
\end{equation}
Note that
\begin{equation}
\label{exp_adv_diff_after_i}
\begin{split}
	&\ \mathbb{E}_{s_{t}\sim\rho_{t}^{\pi'}|n_{i}=0}\left[\overline{A}(s_{t})\right] - \mathbb{E}_{s_{t}\sim\rho_{t}^{\pi}|n_{i}=0}\left[\overline{A}(s_{t})\right] \\
	=&\ \int_{\mathcal{S}}\big[\rho_{t}^{\pi'}(s_{t})-\rho_{t}^{\pi}(s_{t})\big]_{n_{i}=0}\overline{A}(s_{t})ds_{t} \\
	=&\ \int_{\mathcal{S}}\Big(\int_{\mathcal{S}}\big[\rho_{i}^{\pi'}(s_{i})\nu_{\pi'}^{t-i}(s_{t}|s_{i})-\rho_{i}^{\pi}(s_{i})\nu_{\pi}^{t-i}(s_{t}|s_{i})\big]_{{\color{red}n_{i}=0}}ds_{i}\Big)\overline{A}(s_{t})ds_{t} \\
	=&\ \iint_{\mathcal{S}\times\mathcal{S}}\rho_{i}^{\pi}(s_{i})\big[\nu_{\pi'}^{t-i}(s_{t}|s_{i})-\nu_{\pi}^{t-i}(s_{t}|s_{i})\big]\overline{A}(s_{t})ds_{i}ds_{t}.\ \ (\text{since}\ \rho_{i}^{\pi'}=\rho_{i}^{\pi}\ \text{when}\ n_{i}=0) \\
	=&\ \iint_{\mathcal{S}\times\mathcal{S}}\rho_{i}^{\pi}(s_{i})\delta^{t-i}(s_{t}|s_{i})\overline{A}(s_{t})ds_{i}ds_{t}\ \ \ ({\color{blue}\text{denote }\delta^{t-i}(s_{t}|s_{i})=\nu_{\pi'}^{t-i}(s_{t}|s_{i})-\nu_{\pi}^{t-i}(s_{t}|s_{i})}) \\
	=&\ \iint_{\mathcal{S}\times\mathcal{S}}\rho_{i}^{\pi}(s)\delta^{t-i}(s'|s)\overline{A}(s')dsds'
\end{split}
\end{equation}

Substituting \eqref{exp_adv_diff_after_i} into \eqref{exp_adv_diff_upper_bound_almost}, we get
\begin{equation}
\label{exp_adv_diff_upper_bound}
	\mathbb{E}_{s_{t}\sim\rho_{t}^{\pi'}}\left[\overline{A}(s_{t})\right] - \mathbb{E}_{s_{t}\sim\rho_{t}^{\pi}}\left[\overline{A}(s_{t})\right] = \sum_{i=0}^{t-1}P(n_{i}=0)\mathbb{E}_{s_{i}\sim\rho_{i}^{\pi}}\left[P(a'_{i}\neq a_{i}|s_{i})\right]\iint_{\mathcal{S}\times\mathcal{S}}\rho_{i}^{\pi}(s)\delta^{t-i}(s'|s)\overline{A}(s')dsds'.
\end{equation}
For notational simplicity, we denote
\begin{equation}
\label{simplified_ni_prob}
{\color{blue}P_{n_{i}=0}\overline{P}_{a'_{i}\neq a_{i}}:=P(n_{i}=0)\mathbb{E}_{s_{i}\sim\rho_{i}^{\pi}}\left[P(a'_{i}\neq a_{i}|s_{i})\right]}.
\end{equation}
Then, Equation \eqref{exp_adv_diff_upper_bound} can be expressed as
\begin{equation}
\label{simplified_exp_adv_diff_upper_bound}
	\mathbb{E}_{s_{t}\sim\rho_{t}^{\pi'}}\left[\overline{A}(s_{t})\right] - \mathbb{E}_{s_{t}\sim\rho_{t}^{\pi}}\left[\overline{A}(s_{t})\right] = \sum_{i=0}^{t-1}P_{n_{i}=0}\overline{P}_{a'_{i}\neq a_{i}}\iint_{\mathcal{S}\times\mathcal{S}}\rho_{i}^{\pi}(s)\delta^{t-i}(s'|s)\overline{A}(s')dsds'.
\end{equation}

ii) The second part of the proof is given as follows.

Substituting \eqref{simplified_exp_adv_diff_upper_bound} into \eqref{lemma_difference}, we get
\begin{equation}
\label{almost_there}
\begin{split}
	&\ J(\pi') - L_{\pi}(\pi') \\
=&\ \sum_{t=1}^{\infty}\gamma^{t}
\sum_{i=0}^{t-1}P_{n_{i}=0}\overline{P}_{a'_{i}\neq a_{i}}\iint_{\mathcal{S}\times\mathcal{S}}\rho_{i}^{\pi}(s)\delta^{t-i}(s'|s)\overline{A}(s')dsds' \\
\ =&\ \Bigg(P_{n_{0}=0}\overline{P}_{a'_{0}\neq a_{0}}\iint_{\mathcal{S}\times\mathcal{S}}\gamma\rho_{0}^{\pi}(s)\delta^{1}(s'|s)\overline{A}(s')dsds'\Bigg)\ + \\
&\ \Bigg(P_{n_{0}=0}\overline{P}_{a'_{0}\neq a_{0}}\iint_{\mathcal{S}\times\mathcal{S}}\gamma^{2}\rho_{0}^{\pi}(s)\delta^{2}(s'|s)\overline{A}(s')dsds' + P_{n_{1}=0}\overline{P}_{a'_{1}\neq a_{1}}\iint_{\mathcal{S}\times\mathcal{S}}\gamma^{2}\rho_{1}^{\pi}(s)\delta^{1}(s'|s)\overline{A}(s')dsds'\Bigg) +\\
&\ \ \ \vdots \\
\ =&\ P_{n_{0}=0}\overline{P}_{a'_{0}\neq a_{0}}\iint_{\mathcal{S}\times\mathcal{S}}\rho_{0}^{\pi}(s)\big[\gamma\delta^{1}(s'|s)+\gamma^{2}\delta^{2}(s'|s)+\cdots\big]\overline{A}(s')dsds'\ + \\
\ &\ P_{n_{1}=0}\overline{P}_{a'_{1}\neq a_{1}}\iint_{\mathcal{S}\times\mathcal{S}}\gamma\rho_{1}^{\pi}(s)\big[\gamma\delta^{1}(s'|s)+\gamma^{2}\delta^{2}(s'|s)+\cdots\big]\overline{A}(s')dsds'\ + \\
&\ \cdots \\
\ =&\ \frac{1}{1-\gamma}\sum_{t=0}^{\infty}P_{n_{t}=0}\overline{P}_{a'_{t}\neq a_{t}}\iint_{\mathcal{S}\times\mathcal{S}}\gamma^{t}\rho_{t}^{\pi}(s)\big[\mu_{\pi'}(s'|s)-\mu_{\pi}(s'|s)\big]\overline{A}(s')dsds'.\ \ \ (\text{\color{blue}See the definition of $\mu_{\pi}$ in \eqref{mu_definition}.)}
\end{split}
\end{equation}
Taking absolute values on both sides and applying H\"{o}lder's inequality, we get
\begin{equation}
\begin{split}
	\vert J(\pi') - L_{\pi}(\pi')\vert\leq&\  \frac{1}{1-\gamma}\sum_{t=0}^{\infty}P_{n_{t}=0}\overline{P}_{a'_{t}\neq a_{t}}\iint_{\mathcal{S}\times\mathcal{S}}\gamma^{t}\rho_{t}^{\pi}(s)\big\vert\big[\mu_{\pi'}(s'|s)-\mu_{\pi}(s'|s)\big]\overline{A}(s')\big\vert ds'ds \\
	\leq&\ \frac{1}{1-\gamma}\sum_{t=0}^{\infty}P_{n_{t}=0}\overline{P}_{a'_{t}\neq a_{t}}\int_{\mathcal{S}}\gamma^{t}\rho_{t}^{\pi}(s)\int_{\mathcal{S}}\big\vert\mu_{\pi'}(s'|s)-\mu_{\pi}(s'|s)\big\vert ds'ds\cdot \max_{s',a'}|A_{\pi}(s',a')|
\end{split}
\end{equation}
Applying Lemma \ref{lemma2}, we have
\begin{equation}
\label{first_bound}
\begin{split}
	\vert J(\pi') - L_{\pi}(\pi')\vert \leq \frac{2\gamma^{2}\epsilon}{(1-\gamma)^2}\sum_{t=0}^{\infty}P_{n_{t}=0}\overline{P}_{a'_{t}\neq a_{t}}\iint_{\mathcal{S}\times\mathcal{S}}\gamma^{t}\rho_{t}^{\pi}(s)\mu_{\pi}(s'|s)D_{\mathrm{TV}}[\pi'||\pi](s')dsds'.
\end{split}
\end{equation}
Note that the integral part in the above inequality can be expressed as
\begin{equation}
\label{first_avg_DTV}
\begin{split}
	&\ \iint_{\mathcal{S}\times\mathcal{S}}\gamma^{t}\rho_{t}^{\pi}(s)\mu_{\pi}(s'|s)D_{\mathrm{TV}}[\pi'||\pi](s')dsds' \\
	=&\ \int_{\mathcal{S}}\Big(\int_{\mathcal{S}}\gamma^{t}\rho_{t}^{\pi}(s)\mu_{\pi}(s'|s)ds\Big)D_{\mathrm{TV}}[\pi'||\pi](s')ds' \ \ \ (\text{\color{blue}See the definition of $\mu_{\pi}$ in \eqref{mu_definition}.)}\\
	=&\ \int_{\mathcal{S}}\Big(d^{\pi}(s')-(1-\gamma)\sum_{i=0}^{t-1}\gamma^{i}\rho_{i}^{\pi}(s')\Big)D_{\mathrm{TV}}[\pi'||\pi](s')ds'\ \ \ (\text{for all}\ t>0) \\
	=&\ \mathbb{E}_{s'\sim {\color{red}d^{\pi}}}[D_{\mathrm{TV}}[\pi'||\pi](s')]-(1-\gamma)\sum_{i=0}^{t-1}\gamma^{i}\mathbb{E}_{s'\sim {\color{red}\rho_{i}^{\pi}}}[D_{\mathrm{TV}}[\pi'||\pi](s')]\ \ \ (\text{for all}\ t>0) \\
	=&\ \mathbb{E}_{s\sim d^{\pi}}[D_{\mathrm{TV}}[\pi'||\pi](s)]-(1-\gamma)\sum_{i=0}^{t-1}\gamma^{i}\mathbb{E}_{s\sim \rho_{i}^{\pi}}[D_{\mathrm{TV}}[\pi'||\pi](s)]\ \ \ (\text{for all}\ t>0)
\end{split}
\end{equation}

In the following, we will replace all total variations $D_{\mathrm{TV}}[\pi'||\pi](s)$ with $\alpha(s)$ (see Definition \ref{coupling}) and use the following notations for simplicity:
\begin{equation}
\label{alpha}
	\overline{\alpha}:=\mathbb{E}_{s\sim {\color{red}d^{\pi}}}[\alpha(s)],\   \overline{\alpha}_{i}:=\mathbb{E}_{s\sim {\color{red}\rho_{i}^{\pi}}}[\alpha(s)].
\end{equation}
Plugging \eqref{first_avg_DTV} into \eqref{first_bound}, we have
\begin{equation}
\label{second_bound}
\begin{split}
	\vert J(\pi') - L_{\pi}(\pi')\vert \leq\ \frac{2\gamma^{2}\epsilon}{(1-\gamma)^{2}}\Bigg[\sum_{t=0}^{\infty}P_{n_{t}=0}\overline{P}_{a'_{t}\neq a_{t}}\cdot\overline{\alpha}-(1-\gamma)\sum_{t=1}^{\infty}\sum_{i=0}^{t-1}P_{n_{t}=0}\overline{P}_{a'_{t}\neq a_{t}}\cdot\gamma^{i}\overline{\alpha}_{i}\Bigg]
\end{split}
\end{equation}
Using Equations \eqref{disagree_t} and \eqref{simplified_ni_prob}, we have
\begin{equation}
	\sum_{t=1}^{k-1}P_{n_{t}=0}\overline{P}_{a'_{t}\neq a_{t}}=P[n_{k}>0]\rightarrow 1\ \ \text{when}\ k\rightarrow \infty.
\end{equation}
Therefore, the first term in the parentheses on the rightest-hand side of \eqref{second_bound} is just 
\begin{equation}
\label{first_term_of_second_bound}
	\sum_{t=1}^{\infty}P_{n_{t}=0}\overline{P}_{a'_{t}\neq a_{t}}\cdot\overline{\alpha}=\overline{\alpha}.
\end{equation}
The second term in the parentheses on the rightest-hand side of \eqref{second_bound} can be expressed as
\begin{equation}
\label{second_term_of_second_bound}
\begin{split}
	&\ (1-\gamma)\sum_{t=1}^{\infty}\sum_{i=0}^{t-1}P_{n_{t}=0}\overline{P}_{a'_{t}\neq a_{t}}\cdot\gamma^{i}\overline{\alpha}_{i} \\
	=&\ (1-\gamma)\Big(P_{n_{1}=0}\overline{P}_{a'_{1}\neq a_{1}}\cdot\overline{\alpha}_{0}\Big)\ + \\
	&\ (1-\gamma)\Big(P_{n_{2}=0}\overline{P}_{a'_{2}\neq a_{2}}\cdot\overline{\alpha}_{0}+P_{n_{2}=0}\overline{P}_{a'_{2}\neq a_{2}}\cdot\gamma\overline{\alpha}_{1}\Big)\ + \\
	&\ (1-\gamma)\Big(P_{n_{3}=0}\overline{P}_{a'_{3}\neq a_{3}}\cdot\overline{\alpha}_{0}+P_{n_{3}=0}\overline{P}_{a'_{3}\neq a_{3}}\cdot\gamma\overline{\alpha}_{1}\Big)+P_{n_{3}=0}\overline{P}_{a'_{3}\neq a_{3}}\cdot\gamma^{2}\overline{\alpha}_{2}\Big)\ + \\
	&\ \ \ \vdots \\
	=&\ (1-\gamma)\Big(\overline{\alpha}_{0}\sum_{t=1}^{\infty}P_{n_{t}=0}\overline{P}_{a'_{t}\neq a_{t}}+\gamma\overline{\alpha}_{1}\sum_{t=2}^{\infty}P_{n_{t}=0}\overline{P}_{a'_{t}\neq a_{t}}+\gamma^{2}\overline{\alpha}_{2}\sum_{t=3}^{\infty}P_{n_{t}=0}\overline{P}_{a'_{t}\neq a_{t}}+\cdots\Big) \\
	=&\ (1-\gamma)\Big(\overline{\alpha}_{0}[1-P(n_{1}>0)]+\gamma\overline{\alpha}_{1}[1-P(n_{2}>0)]+\gamma^{2}\overline{\alpha}_{2}[1-P(n_{3}>0)]+\cdots\Big) \\
	=&\ (1-\gamma)\Big(\sum_{t=0}^{\infty}\gamma^{t}\overline{\alpha}_{t}-\sum_{t=0}^{\infty}\gamma^{t}\overline{\alpha}_{t}P(n_{t+1}>0)\Big) \\
	=&\ \overline{\alpha} - (1-\gamma)\sum_{t=0}^{\infty}\gamma^{t}\overline{\alpha}_{t}P(n_{t+1}>0)
\end{split}
\end{equation}
Substituting \eqref{first_term_of_second_bound} and \eqref{second_term_of_second_bound} into \eqref{second_bound}, we get
\begin{equation}
\label{second_bound_simplified}
\begin{split}
	\vert J(\pi') - L_{\pi}(\pi')\vert\leq \frac{2\gamma^{2}\epsilon}{1-\gamma}\sum_{t=0}^{\infty}\gamma^{t}\overline{\alpha}_{t}P(n_{t+1}>0).
\end{split}
\end{equation}

iii) The third part of the proof is given as follows.

Recall that $n_t$ denote the number of times that $a'_{i}\neq a_{i}|s_{i}$ at state $s_i$ for $i<t$, and $n_t=0$ means that $a'_{i}=a_{i}|s_{i}$ for all $i<t$. Based on Definition \ref{coupling} (\emph{$\alpha$-coupled policy}), we have $P(a'_{i}=a_{i}|s_{i})\geq 1-\alpha(s_{i})$ for every $s_{i}$. Thus,
\begin{equation}
\label{connect_to_coupling}
\begin{split}
	P(n_{t+1}>0)&\ =1-P(n_{t+1}=0) \\
	&\ =1-\prod_{i=0}^{t}\mathbb{E}_{s_{i}\sim\rho_{i}^{\pi}}\left[P(a'_{i}=a_{i}|s_{i})\right] \\
	&\ \leq 1-\prod_{i=0}^{t}\left(1-\overline{\alpha}_{i}\right)
\end{split}
\end{equation}

Substituting \eqref{connect_to_coupling} and \eqref{alpha} into \eqref{second_bound_simplified}, we have
\begin{equation}
\label{third_bound}
	\vert J(\pi') - L_{\pi}(\pi')\vert\leq \frac{2\gamma^{2}\epsilon}{1-\gamma}\sum_{t=0}^{\infty}\gamma^{t}\overline{\alpha}_{t}\Big(1-\prod_{i=0}^{t}\left(1-\overline{\alpha}_{i}\right)\Big).
\end{equation}

Using Lemma \ref{lemma3}, the inequality \eqref{third_bound} can be further simplified as
\begin{equation}
	\vert J(\pi') - L_{\pi}(\pi')\vert\leq \frac{2\gamma^{2}\epsilon}{(1-\gamma)^{3}}\sum_{t=0}^{\infty}\gamma^{t}\overline{\alpha}_{t}^{2}.
\end{equation}
Replacing $\overline{\alpha}_{t}$ with $\mathbb{E}_{s\sim \rho_{t}^{\pi}}[D_{\mathrm{TV}}[\pi'||\pi](s)]$ and applying $\mathbb{E}^{2}[X]\leq\mathbb{E}[X^{2}]$, we get
\begin{equation}
\label{meaningful_bound}
	\vert J(\pi') - L_{\pi}(\pi')\vert\leq \frac{2\gamma^{2}\epsilon}{(1-\gamma)^{3}}\sum_{t=0}^{\infty}\gamma^{t}\mathbb{E}_{s\sim \rho_{t}^{\pi}}[D_{\mathrm{TV}}^{2}[\pi'||\pi](s)].
\end{equation}
Last, applying Pinsker's inequality, $2D^{2}_{\mathrm{TV}}[\pi'||\pi](s)\leq D_{\mathrm{KL}}[\pi'||\pi](s)$, the result follows.
\end{proof}

\section{Proof of Analytical Policy Update Rule with Monotonic Improvement Guarantee }
\label{appendix-B}
This proof uses \emph{calculus of variations} to derive an analytical solution for trust region policy update.
\begin{theorem}
\label{formula-appendix}
For any stochastic policies $\pi_{\mathrm{new}},\pi_{\mathrm{old}}$ that are continuously differentiable on the state space $\mathcal{S}$, the inequality, $J(\pi_{\mathrm{new}})\geq J(\pi_{\mathrm{old}})$, holds when
\begin{equation}
\label{proof_update_formula}
\pi_{\mathrm{new}}=\pi_{\mathrm{old}}\cdot\frac{e^{\alpha_{\pi_{\mathrm{old}}}}}{\mathbb{E}_{a\sim\pi_{\mathrm{old}}}\left[e^{\alpha_{\pi_{\mathrm{old}}}}\right]},\ \text{where } \alpha_{\pi_{\mathrm{old}}}=\frac{A_{\pi_{\mathrm{old}}}}{C_{\pi_{\mathrm{old}}}}.
\end{equation}
\end{theorem}

\begin{proof}
With Theorem \ref{new_bound}, we can get a lower bound of the objective function $J(\pi')$ when approximating around $\pi_{old}$:
\begin{equation}
	J(\pi') \geq L_{\pi_{\mathrm{old}}}(\pi') - \frac{1}{1-\gamma}C_{\pi_{\mathrm{old}}}\mathbb{E}_{s\sim d^{\pi_{\mathrm{old}}}}\left[D_{\mathrm{KL}}[\pi'\Vert\pi_{\mathrm{old}}](s)\right].
\end{equation}
 It follows that maximizing the lower bound will give us a new policy that is not worse than $\pi_{\mathrm{old}}$. To see this, let $I(\pi')$ denote the lower bound and $\pi_{\mathrm{new}}$ denote its maximum solution:
\begin{equation}
	I(\pi') = \ L_{\pi_{\mathrm{old}}}(\pi')-\frac{1}{1-\gamma}C_{\pi_{\mathrm{old}}}\mathbb{E}_{s\sim d^{\pi_{\mathrm{old}}}}\left[D_{\mathrm{KL}}[\pi'\Vert\pi_{\mathrm{old}}](s)\right]
\end{equation}
\begin{equation}
\label{max}
	\pi_{\mathrm{new}} = \argmax_{\pi'}I(\pi')
\end{equation}
where $L_{\pi_{\mathrm{old}}}(\pi')$ is the surrogate model. Then, we have 
$$J(\pi_{\mathrm{new}})\geq I(\pi_{\mathrm{new}})\geq I(\pi_{\mathrm{old}})=J(\pi_{\mathrm{old}}).$$
Next, we prove that the expression of $\pi_{\mathrm{new}}$ in \eqref{proof_update_formula} is a necessary and sufficient condition for the optimal solution of the problem in \eqref{max}.

\subsection{Continuous action space}
We will use \emph{calculus of variation} to derive the analytical expression for $\pi_{\mathrm{new}}$. Let $\pi'\in C^{1}(U)$ be functions defined on $U\doteq\mathcal{S}\times\mathcal{A}$. Note that the lower bound $I(\pi')$ can be rewritten as follows:
\begin{equation}
\label{functional_obj}
	I(\pi')=J(\pi_{\mathrm{old}})+\frac{1}{1-\gamma}\iint_{\mathcal{S}\times\mathcal{A}}d^{\pi_{\mathrm{old}}}(s)\Bigg[\pi'(a|s)A_{\pi_{\mathrm{old}}}(s,a)-C_{\pi_{\mathrm{old}}}\pi'(a|s)\log\frac{\pi'(a|s)}{\pi_{\mathrm{old}}(a|s)}\Bigg] ds da.
\end{equation}
Note that the policy $\pi'$ should be a probability distribution, which means that it integrates to 1. To ensure that, we add the following constraint:
\begin{equation}
\label{functional_cons}
	H(\pi')=\frac{1}{1-\gamma}\int_{\mathcal{S}}d^{\pi_{\mathrm{old}}}(s) \Big[\int_{\mathcal{A}}\pi'(a|s) da - 1\Big] ds=0.
\end{equation}
Now, consider all functions in Equations \eqref{functional_obj} and \eqref{functional_cons} as variables in function spaces, and define
\begin{equation}
\label{lagrange_obj}
	F(s,a,\pi')=d^{\pi_{\mathrm{old}}}\left(\pi'A_{\pi_{\mathrm{old}}}-C_{\pi_{\mathrm{old}}}\pi'\log\pi'+C_{\pi_{\mathrm{old}}}\pi'\log\pi_{\mathrm{old}}\right).
\end{equation}
\begin{equation}
\label{lagrange_cons}
	G(s,a,\pi')=d^{\pi_{\mathrm{old}}}\pi'-d^{\pi_{\mathrm{old}}}\pi_{\mathrm{old}}.\ \ \ \ \ \ \ \ \ \ \ \ \ \ \ \ \ \ \ \ \ \ \ \ \ \ \ \ \ \ \ \ \ \ \ \ \ \ \ \ \ \ \ \ \ \ \ \ \ \ \ 
\end{equation}
Based on Euler-Lagrange equation \cite{noauthororeditor}, there must exist a real number $\lambda$ such that the optimal policy $\pi^{*}$ satisfies
\begin{equation}
\label{Euler-Lagrange}
	\nabla_{\pi'}F(s,a,\pi')-\lambda\nabla_{\pi'}G(s,a,\pi')=0,
\end{equation}
where $\lambda$ is the Lagrange multiplier. Solving Equation \eqref{Euler-Lagrange}, we have
\begin{equation}
	d^{\pi_{\mathrm{old}}}\left(A_{\pi_{\mathrm{old}}}-C_{\pi_{\mathrm{old}}}\log\pi^{*}-C_{\pi_{\mathrm{old}}}+C_{\pi_{\mathrm{old}}}\log\pi_{\mathrm{old}}-\lambda\right)=0,
\end{equation}
and
\begin{equation}
\label{update_formula_Lagrange}
	\pi_{\mathrm{new}}=\pi_{\mathrm{old}}\cdot\exp\big\{\frac{A_{\pi_{\mathrm{old}}}}{C_{\pi_{\mathrm{old}}}}-1-\frac{\lambda}{C_{\pi_{\mathrm{old}}}}\big\}.
\end{equation}
Since $\pi'$ integrates to 1, we have
\begin{equation}
\begin{split}
	\int_{\mathcal{A}}\pi_{\mathrm{new}}(a|s)da =\ &\ \int_{\mathcal{A}}\pi_{\mathrm{old}}(a|s)\exp\big\{\frac{A_{\pi_{\mathrm{old}}}(s,a)}{C_{\pi_{\mathrm{old}}}}-1-\frac{\lambda}{C_{\pi_{\mathrm{old}}}}\big\} da \\
	=\ &\ e^{-1-\lambda/C_{\pi_{\mathrm{old}}}}\int_{\mathcal{A}}\pi_{\mathrm{old}}(a|s) \exp\big\{\frac{A_{\pi_{\mathrm{old}}}(s,a)}{C_{\pi_{\mathrm{old}}}}\big\} da \\
	=\ &\ 1
\end{split}
\end{equation}
Rearranging it, we get
\begin{equation}
	\int_{\mathcal{A}}\pi_{\mathrm{old}}(a|s) \exp\big\{\frac{A_{\pi_{\mathrm{old}}}(s,a)}{C_{\pi_{\mathrm{old}}}}\big\} da = e^{1+\lambda/C_{\pi_{\mathrm{old}}}}.
\end{equation}
Taking logarithm on both sides and rearranging it, we get
\begin{equation}
\label{lambda}
	\lambda = C_{\pi_{\mathrm{old}}}\log\int_{\mathcal{A}}\pi_{\mathrm{old}}(a|s) \exp\big\{\frac{A_{\pi_{\mathrm{old}}}(s,a)}{C_{\pi_{\mathrm{old}}}}\big\} da - C_{\pi_{\mathrm{old}}}.
\end{equation}
Substituting \eqref{lambda} into \eqref{update_formula_Lagrange}, we get
\begin{equation}
	\pi_{\mathrm{new}}(a|s)=\pi_{\mathrm{old}}\cdot\exp\big\{\frac{A_{\pi_{\mathrm{old}}}(s,a)}{C_{\pi_{\mathrm{old}}}}-\log\int_{\mathcal{A}}\pi_{\mathrm{old}}(a|s) \exp\big\{\frac{A_{\pi_{\mathrm{old}}}(s,a)}{C_{\pi_{\mathrm{old}}}}\big\} da\big\}.
\end{equation}
Denote $\alpha_{\pi_{\mathrm{old}}}=\frac{A_{\pi_{\mathrm{old}}}(s,a)}{C_{\pi_{\mathrm{old}}}}$. Then, the optimal policy can be simplified as
\begin{equation}
\label{sufficient_condition}
	\pi_{\mathrm{new}}=\pi_{\mathrm{old}}\cdot\frac{e^{\alpha_{\pi_{\mathrm{old}}}}}{\mathbb{E}_{a\sim\pi_{\mathrm{old}}}\left[e^{\alpha_{\pi_{\mathrm{old}}}}\right]}.
\end{equation}

Until now, we have proved the sufficient condition. Next, we prove that the policy $\pi_{\mathrm{new}}$ in Eq. (\ref{sufficient_condition}) is also the necessary condition for the optimal solution to the maximization of $I(\pi')$. 

Consider weak variations $\epsilon\eta$ such that $\pi'=\pi_{\mathrm{new}}+\epsilon\eta$, where $\eta\in C^{1}(\overline{U})$ and $\epsilon$ is a real number. The second variation can be expressed as,
\begin{equation}
\label{second-variation}
\begin{split}
	\delta^{2}I &=\ \frac{\epsilon^{2}}{1-\gamma}\iint_{\mathcal{S}\times\mathcal{A}}\nabla^{2}_{\pi'\pi'}F(s,a,\pi')\eta^{2} dsda \\
	&=\ \frac{\epsilon^{2}}{1-\gamma}\iint_{\mathcal{S}\times\mathcal{A}} -\frac{d^{\pi_{\mathrm{old}}}C_{\pi_{\mathrm{old}}}}{\pi'}\eta^{2} dsda \\
	&\leq\ 0\ \ \ \ \ \ \text{(for all weak variations $\eta$)}
\end{split}
\end{equation}
because $C_{\pi_{\mathrm{old}}}\geq 0$, and $d^{\pi_{\mathrm{old}}},\pi'$ are probability distributions and thus always greater or equal to 0. Based on second-variation condition \cite{MarkKot2014}, the functional $I(\pi')$ reaches a maximum at $\pi_{\mathrm{new}}$.

\subsection{Discrete action space}
For discrete actions, the functionals $I(\pi')$ and $H(\pi')$ can be rewritten as follows:
\begin{equation}
\label{functional_obj_discrete}
	I(\pi')=J(\pi_{\mathrm{old}})+\frac{1}{1-\gamma}\int_{\mathcal{S}}d^{\pi_{\mathrm{old}}}(s)\sum_{i=1}^{k}\Bigg[\pi'(a_{i}|s)A_{\pi_{\mathrm{old}}}(s,a_{i})-C_{\pi_{\mathrm{old}}}\pi'(a_{i}|s)\log\frac{\pi'(a_{i}|s)}{\pi_{\mathrm{old}}(a_{i}|s)}\Bigg] ds,
\end{equation}
\begin{equation}
\label{functional_cons_discrete}
	H(\pi')=\frac{1 }{1-\gamma}\int_{\mathcal{S}}d^{\pi_{\mathrm{old}}}(s) \Big[\sum_{i=1}^{k} \pi'(a_{i}|s) - 1\Big] ds.
\end{equation}
Now, consider the policy as a vector of functions, $\pi'=[\pi'_{1},\pi'_{2},\dots,\pi'_{k}]$, where $\pi'_{i}=\pi'(a_{i}|s)\in C^{1}(\mathcal{S})$ is a function defined on $\mathcal{S}$ given the action $a_{i}$. Then, we can define the Lagrange functions by
\begin{equation}
\label{lagrange_obj_discrete}
	F(s,a,\pi')=d^{\pi_{\mathrm{old}}}\sum_{i=1}^{k}\Big(\pi'_{i}A_{\pi_{\mathrm{old}}}-C_{\pi_{\mathrm{old}}}\pi'_{i}\log\pi'_{i}+C_{\pi_{\mathrm{old}}}\pi'_{i}\log\pi_{i,\mathrm{old}}\Big).
\end{equation}
\begin{equation}
\label{lagrange_cons}
	G(s,a,\pi')=\sum_{i=1}^{k} \pi'_{i} - 1.\ \ \ \ \ \ \ \ \ \ \ \ \ \ \ \ \ \ \ \ \ \ \ \ \ \ \ \ \ \ \ \ \ \ \ \ \ \ \ \ \ \ \ \ \ \ \ \ \ \ \ \ \ \ \ \ \ \ \ \ \ \ \ \ \ \ \ \ \ \ \ \ \ \ \ 
\end{equation}
The Euler-Lagrange Equation \eqref{Euler-Lagrange} becomes
\begin{equation}
\label{Euler-Lagrange-discrete}
	\nabla_{\pi'_{i}}F(s,a,\pi')-\lambda\nabla_{\pi'_{i}}G(s,a,\pi')=0,\ \forall i\in\{1,\dots,k\}.
\end{equation}
Solving the Euler-Lagrange Equations \eqref{Euler-Lagrange-discrete}, we get
\begin{equation}
\label{update_formula_Lagrange_discrete}
	\pi^{*}_{i}=\pi_{i,\mathrm{old}}\cdot\exp\big\{\frac{A_{\pi_{\mathrm{old}}}}{C_{\pi_{\mathrm{old}}}}-1-\frac{\lambda}{C_{\pi_{\mathrm{old}}}}\big\},\ \forall i\in\{1,\dots,k\}.
\end{equation}
Note that $\pi^{*}_{i}$ should satisfy
\begin{equation}
	\sum_{i=1}^{k}\pi^{*}_{i}=1.
\end{equation}
Then, we can calculate the Lagrange multiplier $\lambda$:
\begin{equation}
\label{lambda_discrete}
	\lambda = C_{\pi_{\mathrm{old}}}\log\sum_{i=1}^{k} \pi_{i,\mathrm{old}}(a_{i}|s) \exp\big\{\frac{A_{\pi_{\mathrm{old}}}(s,a_{i})}{C_{\pi_{\mathrm{old}}}}\big\} - C_{\pi_{\mathrm{old}}}.
\end{equation}
Substituting \eqref{lambda_discrete} into \eqref{update_formula_Lagrange_discrete} and use the vector form, we get
\begin{equation}
\label{sufficient_condition_discrete}
	\pi^{*}=\pi_{\mathrm{new}}=\pi_{\mathrm{old}}\cdot\frac{e^{\alpha_{\pi_{\mathrm{old}}}}}{\mathbb{E}_{a\sim\pi_{\mathrm{old}}}\left[e^{\alpha_{\pi_{\mathrm{old}}}}\right]}.
\end{equation}
Use the same method as in Equation \eqref{second-variation}, we can prove that the second-variation condition is satisfied.
\end{proof}

%%%%%%%%%%%%%%%%%%%%%%%%%%%%%%%%%%%%%%%%%%%%%%%%%%%%%%%%%%%%%%%%%%%%%%%%%%%%%%%
%%%%%%%%%%%%%%%%%%%%%%%%%%%%%%%%%%%%%%%%%%%%%%%%%%%%%%%%%%%%%%%%%%%%%%%%%%%%%%%

\end{document}